\documentclass[sigconf]{acmart}
\AtBeginDocument{%
  }

\copyrightyear{2026}
\acmYear{2026}
\setcopyright{cc}
\setcctype{by}
\acmConference[ICMI '26]{INTERNATIONAL CONFERENCE ON MULTIMODAL INTERACTION}{October 05--09, 2026}{Napoli, Italy}
\acmBooktitle{INTERNATIONAL CONFERENCE ON MULTIMODAL INTERACTION (ICMI '26), October 05--09, 2026, Napoli, Italy}
\acmDOI{10.1145/3776574.3831225}
\acmISBN{979-8-4007-2318-6/2026/10}

\usepackage{multirow}
\usepackage{makecell}
\usepackage{subcaption}

\begin{document}

\title{SENSE: Efficient EEG-to-Text via Privacy-Preserving Semantic Retrieval}


\author{Akshaj Murhekar}
\authornote{Both authors contributed equally to this research.}
\email{akshaj.murhekar@utexas.edu}
\affiliation{%
  \institution{The University of Texas at Austin}
  \department{School of Information}
  \city{Austin}
  \state{Texas}
  \country{USA}
}

\author{Christina Liu}
\authornotemark[1]
\email{cl49786@utexas.edu}
\affiliation{%
  \institution{The University of Texas at Austin}
  \department{School of Information}
  \city{Austin}
  \state{Texas}
  \country{USA}
}

\author{Abhijit Mishra}
\email{abhijitmishra@utexas.edu}
\affiliation{%
  \institution{The University of Texas at Austin}
  \department{School of Information}
  \city{Austin}
  \state{Texas}
  \country{USA}
}

\author{Shounak Roychowdhury}
\email{shounak.roychowdhury@ischool.utexas.edu}
\affiliation{%
  \institution{The University of Texas at Austin}
  \department{School of Information}
  \city{Austin}
  \state{Texas}
  \country{USA}
}

\author{Jacek Gwizdka}
\email{jacekg@utexas.edu}
\affiliation{%
 \institution{School of Information\\The University of Texas at Austin}
  \city{}
  \country{United States}}
\affiliation{%
 \institution{Institute of Applied Computer Science\\ Łódź University of Technology}
  \city{}
  \country{Poland}}

\renewcommand{\shortauthors}{Murhekar and Liu, et al.}



\begin{abstract}
Decoding brain activity into natural language is a major challenge in AI with important applications in assistive communication, neurotechnology, and human--computer interaction. Most existing Brain--Computer Interface (BCI) approaches rely on memory-intensive fine-tuning of Large Language Models (LLMs) or encoder--decoder models on raw EEG signals, resulting in expensive training pipelines, limited accessibility, and potential exposure of sensitive neural data. We introduce \textsc{SENSE} (\textbf{SE}mantic \textbf{N}eural \textbf{S}parse \textbf{E}xtraction), a lightweight and privacy-preserving framework that translates non-invasive electroencephalography (EEG) into text without LLM fine-tuning. \textsc{SENSE} decouples decoding into two stages: on-device semantic retrieval and prompt-based language generation. EEG signals are locally mapped to a discrete textual space to extract a non-sensitive Bag-of-Words (BoW), which conditions an off-the-shelf LLM to synthesize fluent text in a zero-shot manner. The EEG-to-keyword module contains only $\sim$6M parameters and runs fully on-device, ensuring raw neural signals remain local while only abstract semantic cues interact with language models. Evaluated on a 128-channel EEG dataset across six subjects, \textsc{SENSE} matches or surpasses the generative quality of fully fine-tuned baselines such as \textsc{Thought2Text} while substantially reducing computational overhead. By localizing neural decoding and sharing only derived textual cues, \textsc{SENSE} provides a scalable and privacy-aware retrieval-augmented architecture for next-generation BCIs.
\end{abstract}



  


\maketitle

\section{Introduction}
\label{sec:intro}

\begin{figure}[t]
    \centering
    \Description{A vertical flowchart illustrating the SENSE privacy-preserving data flow. At the top, a brain icon emits Raw EEG signals. These signals enter a box labeled YOUR LOCAL DEVICE (ON-DEVICE), which contains a dashed Privacy Preserving Boundary. Inside this boundary, an 'EEG ENCODER' generates a Pooled EEG (512-d) vector, which is then processed by the Similarity Refiner neural network. The refiner outputs a Non-sensitive Bag-of-words (BoW) —example keywords shown are backrest, chair, cushion, table. This discrete BoW is the only information that exits the local device to reach a Remote LLM Decoder (Gemini, ChatGPT, LLaMa). At the bottom, the remote LLM produces the final output: "A wooden bench with a cushion and backrest."}
    \includegraphics[width=0.67\columnwidth]{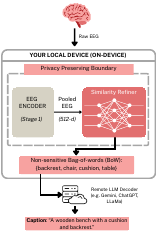} 
    \caption{The SENSE privacy-preserving architecture. Raw, high-dimensional EEG signals are processed within a local "Privacy Preserving Boundary" to extract non-sensitive semantic anchors. Only discrete, interpretable keywords are transmitted to remote Large Language Models for final linguistic synthesis.}
    \label{fig:privacy_teaser}
\end{figure}

Translating human thoughts into multimodal artifacts such as text and images has long been a central objective in neurotechnology and artificial intelligence, with recent breakthroughs in generative AI accelerating research in brain-signal-to-artifact generation \cite{Speier_2016, benchetrit2023brain, defossez2023decoding}. In this work, we focus on brain-signal-to-text generation for \textbf{visual thought to text translation} using non-invasive \textit{electroencephalography} (EEG), one of the most widely used modalities in Brain--Computer Interface (BCI) systems due to its portability, safety, and affordability \cite{he2015noninvasive}. Advances in decoding EEG into language and images \cite{Speier_2016, benchetrit2023brain, defossez2023decoding} enable applications such as assistive communication for patients with neurological disorders, immersive AR/VR experiences, mental health monitoring, and interactive gaming. At the same time, progress in Natural Language Processing driven by Large Language Models (LLMs), including GPT \cite{achiam2023gpt}, Gemini \cite{team2023gemini}, LLaMA \cite{touvron2023llama}, Mistral \cite{jiang2023mistral}, and Phi \cite{gunasekar2023textbooks}, has enabled flexible multimodal reasoning across modalities such as vision \cite{liu2024visual} and speech \cite{rubenstein2023audiopalm, fathullah2024prompting}. These developments motivate further research into decoding brain signals into structured language representations.

However, most existing approaches that combine EEG with large language models rely on fine-tuning large neural models directly on brain signals, including autoregressive LMs \cite{mishra-etal-2025-thought2text} and encoder--decoder architectures \cite{wang2024enhancing,lamprou2025creating,tao2025see}. While effective, these methods require substantial computational resources and infrastructure, limiting accessibility and scalability, and transmitting raw EEG signals to external services raises privacy concerns due to the sensitive nature of neural data. Moreover, tightly coupling decoding pipelines to specific models makes them difficult to adapt as generative AI rapidly evolves, with increasingly powerful LLMs and reasoning models released each year. These challenges motivate EEG-to-text frameworks that are lightweight, privacy-aware, and modular enough to scale with advances in generative AI.

In this work we propose a lightweight framework for decoding EEG signals into text without fine-tuning large language models. As illustrated in Figure \ref{fig:privacy_teaser}, our method follows a modular pipeline in which \textbf{EEG signals are mapped to a representation aligned with visual semantics, salient textual cues are extracted, and language models generate the final sentence}. We train a multichannel EEG encoder whose embeddings align with pooled representations from a pretrained CLIP visual encoder \cite{pmlr-v139-radford21a}, and project them into the CLIP text embedding space to infer likely words from a fixed vocabulary via similarity scoring. These keywords serve as structured prompts to LLMs that generate natural language descriptions at inference time, following the pipeline: \textit{EEG $\rightarrow$ CLIP-aligned embedding $\rightarrow$ keyword extraction $\rightarrow$ LLM prompt $\rightarrow$ sentence generation}. LLMs are therefore used only through prompting, requiring no model fine-tuning.

This design provides two practical benefits. The EEG-to-keyword extraction module, including the encoder and projection layers, contains roughly $\sim$6M parameters and can be deployed on-premises with modest resources. In addition, raw EEG signals remain entirely local while only extracted textual keywords may optionally interact with external LLM APIs, enabling privacy-preserving deployments or fully local inference when LLMs are self-hosted.

We evaluate the approach on a public 128-channel EEG dataset collected from six participants viewing visual stimuli \cite{spampinato2017deep}. The dataset pairs images with captions, allowing EEG-to-text modeling under language-agnostic stimulus conditions while emphasizing perceptual representations encoded in neural activity. A fixed vocabulary is constructed from the captions, and keyword-driven generation is evaluated with both proprietary and open models, including ChatGPT-4o-mini \cite{openai2024gpt4ocard}, Gemini 2.5 Flash Lite \cite{team2023gemini}, \textsc{LLaMA3} \cite{grattafiori2024llama3herdmodels}, and \textsc{Qwen2.5-7B} \cite{qwen2025qwen25technicalreport}. Results show that prompt-based generation from EEG-derived keywords achieves quality comparable to, and in some cases exceeding, approaches that rely on fine-tuned LLMs such as \textsc{Thought2Text} \cite{mishra-etal-2025-thought2text}. Such deployable EEG-to-text systems could support applications including assistive communication, everyday lightweight BCI interfaces, and brain-driven interaction in augmented and virtual reality environments. To address these challenges, our key contributions are as follows:
\begin{itemize}
    \item A lightweight EEG-to-text framework that avoids fine-tuning large language models by decomposing the task into EEG-to-keyword extraction followed by prompt-based text generation.
    \item A CLIP-aligned EEG representation that enables semantic grounding of brain signals and supports vocabulary-level keyword inference through similarity in CLIP's text embedding space.
    \item A privacy-aware architecture in which EEG signals remain on-premises while only extracted textual keywords interact with language models.
    \item Empirical evaluation showing that prompt-driven generation using EEG-derived keywords achieves comparable or improved performance relative to LLM fine-tuning approaches such as \textsc{Thought2Text} \cite{mishra-etal-2025-thought2text}.
\end{itemize}

The code has been released for academic purposes at \url{https://github.com/akshajmurhekar/SENSE-EEG}
\section{Related Work}
The pursuit of BCIs for natural language decoding has seen remarkable acceleration over the past five years, shifting from isolated command classification to continuous text generation. Initial breakthroughs in high-fidelity brain-to-text systems predominately relied on invasive neuroprostheses, successfully enabling individuals with severe motor impairments to synthesize speech and text by decoding cortical activity associated with attempted handwriting or vocalization \cite{willett2023high, metzger2022generalizable}. Building on these invasive successes, landmark studies have demonstrated that non-invasive signals can successfully reconstruct perceived speech \cite{defossez2023decoding}, visual stimuli \cite{benchetrit2023brain}, and even continuous semantic narratives \cite{tang2023semantic}, with frameworks like \textsc{Brain2Qwerty} expanding this scope to decode typing processes directly from non-invasive recordings \cite{levy2025brain}. However, extracting granular linguistic information from the inherently noisy and low-spatial-resolution domain of EEG remains a significant hurdle.
\begin{figure*}[t]
    \centering
    \includegraphics[width=2.0\columnwidth]{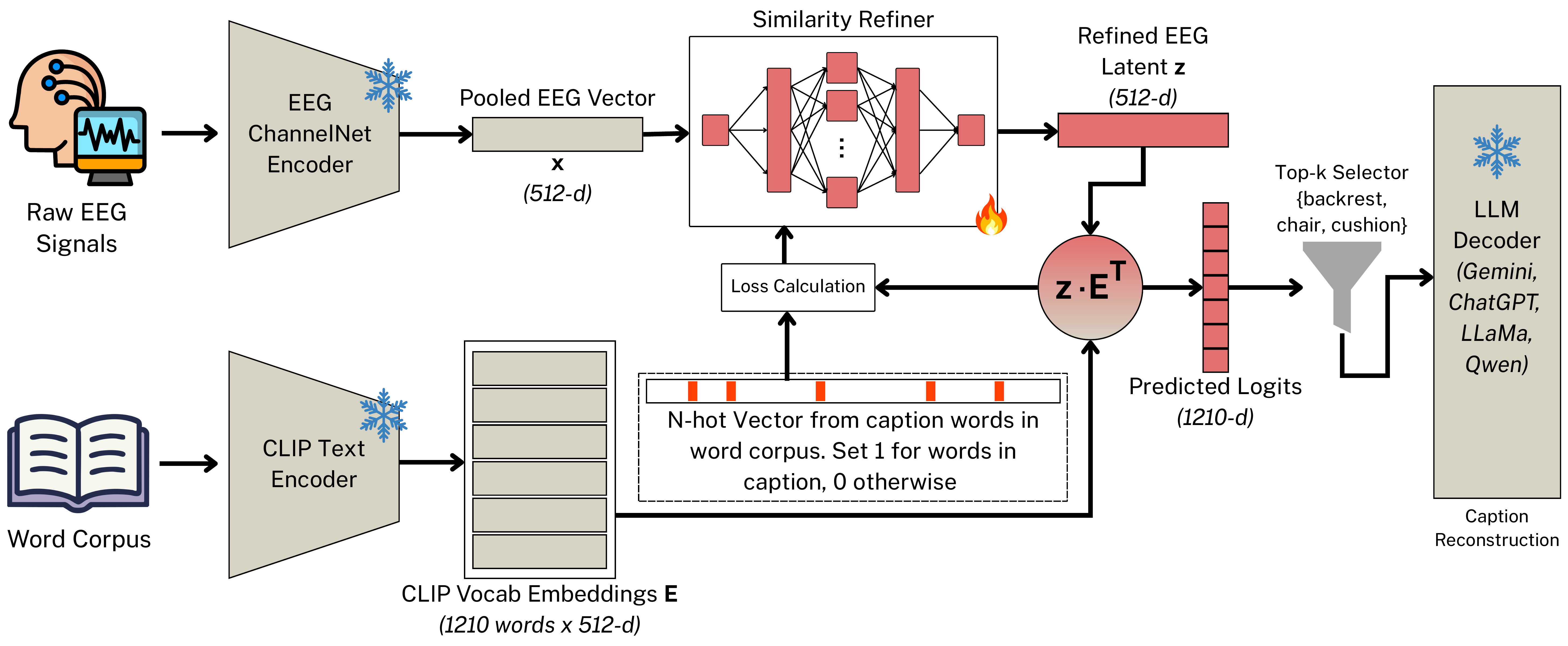}
    \Description{A high-level architectural flowchart of the SENSE pipeline. 
    The diagram is organized into three horizontal stages: Encoding, Refining/Matching, and Decoding. In the Encoding stage, Raw EEG Signals are processed by a frozen EEG ChannelNet Encoder to produce a 512-dimensional vector x. Simultaneously, a Word Corpus is processed by a frozen CLIP Text Encoder to generate a 1210-word vocabulary embedding matrix E. In the Refining stage, vector x enters a Similarity Refiner (indicated as the only trainable component) to produce a refined latent z. This latent z is matrix-multiplied with the transpose of the embedding matrix E to produce predicted logits. A loss calculation compares these logits against an N-hot vector derived from ground-truth caption words to update the refiner. In the Decoding stage, the predicted logits pass through a Top-k Selector funnel, which extracts a Bag-of-Words (e.g., backrest, chair, cushion). This BoW then prompts a frozen LLM Decoder (Gemini, ChatGPT, etc.) to perform the final Caption Reconstruction.}
    
    \caption{\textsc{SENSE} pipeline overview: ChannelNet EEG encoder \cite{mishra-etal-2025-thought2text} extracts vector $\mathbf{x}$; a \textit{Similarity Refiner} maps $\mathbf{x}\!\to\!\mathbf{z}$. $\mathbf{z}$ is matrix-multiplied with CLIP-space vocabulary embeddings $\mathbf{E}$ to produce logits. A top-$k$ selector ($k=15$) yields salient tokens forming a Bag-of-Words (BoW). The BoW and predicted object label prompt an off-the-shelf LLM for caption reconstruction. The refiner is trained against ground-truth $N$-hot vectors with specialized losses addressing class imbalance.}
    \label{fig:Arch}
\end{figure*}
To overcome these signal limitations, contemporary non-invasive text decoding frameworks heavily leverage semi-supervised pre-training and complex cross-modal alignments. Recent works have explored transferable representations using pre-trained contrastive masked autoencoders \cite{wang2024enhancing}, scaling up generalized feature extraction through massive unlabeled datasets \cite{lamprou2025creating}, and enforcing interpretable representation learning to preserve semantic faithfulness \cite{liu2025learning}. To map these signals to language, models often anchor continuous EEG features to pre-trained text spaces using specific matching modules, as seen in SEE \cite{tao2025see} and open-vocabulary approaches like ETS \cite{masry2025ets}. However, these paradigms, including state-of-the-art system like \textsc{Thought2Text} \cite{mishra-etal-2025-thought2text}, which is most related to our work require memory-intensive, end-to-end fine-tuning of autoregressive or encoder-decoder LLMs directly on raw neural data. In contrast, our framework completely bypasses LLM fine-tuning. By framing decoding as a multi-label latent alignment problem rather than a generation task, \textsc{SENSE} achieves competitive text synthesis through lightweight, discrete retrieval rather than heavy model adaptation.

\section{Method}
\label{sec:method}
Our goal is to decode natural language descriptions from EEG signals without relying on end-to-end LLM fine-tuning. Following the task formulation introduced in \textsc{Thought2Text} \cite{mishra-etal-2025-thought2text}, we consider paired examples of EEG signals and textual descriptions associated with visual stimuli. Using visual stimuli instead of textual prompts avoids the complexities of language processing in the brain (e.g., syntactic parsing and sequential word integration) and instead elicits neural responses to salient perceptual features. This enables the learning of language-agnostic neural representations that can later be expressed in natural language. Formally, each sample consists of an EEG recording $\mathbf{X} \in \mathbb{R}^{C \times T}$, where $C$ denotes the number of channels and $T$ the temporal samples, paired with a caption $\mathbf{s}$ describing the viewed image. The objective is to learn a mapping $f: \mathbf{X} \rightarrow \mathbf{s}$ that generates a natural language description capturing the semantic content perceived by the subject.

\textsc{SENSE} translates continuous EEG signals into natural language by decoupling neural feature extraction from linguistic synthesis. The framework (shown in Figure \ref{fig:Arch}) first maps vision-aligned EEG embeddings into a discrete semantic space to retrieve a salient Bag-of-Words (BoW). This intermediate representation provides a compact semantic interface between neural signals and language generation while reducing the burden of directly decoding full sentences from noisy EEG signals. The retrieved concepts are then used to condition an off-the-shelf LLM, which reconstructs fluent text via zero-shot prompting.

\subsection{Semantic Target Formulation}
We first define a global vocabulary $\mathcal{V}$ ($V=1210$) consisting of unique nouns, verbs, and adjectives from ImageNet training captions, lemmatized using WordNet lemmatizer \cite{miller1995wordnet} to their base forms. This vocabulary provides a semantically grounded concept space that captures salient visual attributes while remaining tractable for neural decoding. For each EEG sample $i$, we construct an N-hot target vector $\mathbf{y}_i \in \{0, 1\}^V$, where $y_{i, j} = 1$ if the $j$-th concept is present in the caption. This results in a highly sparse space with an average of 5 active tokens (0.4\%) per sample. We therefore formulate the problem as a multi-label latent alignment task in which the model learns to geometrically map EEG signals to the embeddings of semantically relevant words.

\subsection{Cross-Modal Projector Architecture}
To bridge the gap between vision-aligned EEG embeddings $\mathbf{x} \in \mathbb{R}^{512}$ and the CLIP text embedding space, we introduce a \textit{Similarity Refiner}. A naive zero-shot retrieval baseline using $L_2$-normalized raw logits 
\[
\mathbf{\hat{y}}_{\text{naive}} = \left(\frac{\mathbf{x}}{\|\mathbf{x}\|_2}\right)\left(\frac{\mathbf{E}}{\|\mathbf{E}\|_2}\right)^\top
\]
proves insufficient due to modality noise and the distributional mismatch between neural features and textual embeddings. 

The \textit{Similarity Refiner} (an MLP bottleneck) transforms $\mathbf{x}$ into a refined latent representation $\mathbf{z}$ via:

\begin{equation}
\mathbf{h} = \text{ReLU}(\text{LayerNorm}(\mathbf{W}_1 \mathbf{x} + \mathbf{b}_1))
\end{equation}

\begin{equation}
\mathbf{z} = \mathbf{W}_2 \mathbf{h} + \mathbf{b}_2
\end{equation}

where $\mathbf{W}_1 \in \mathbb{R}^{1024 \times 512}$ and $\mathbf{W}_2 \in \mathbb{R}^{512 \times 1024}$. This bottleneck structure allows the model to refine noisy EEG embeddings while preserving their semantic structure. We enforce semantic alignment between $\mathbf{z}$ and the frozen Vocabulary Matrix $\mathbf{E} \in \mathbb{R}^{V \times 512}$ using cosine similarity. The final predicted logits $\mathbf{\hat{y}} \in \mathbb{R}^V$ are computed as:

\begin{equation}
\mathbf{\hat{y}} =
\left(
\frac{\mathbf{z}}{\|\mathbf{z}\|_2}
\right)
\left(
\frac{\mathbf{E}}{\|\mathbf{E}\|_2}
\right)^\top
\end{equation}

This formulation interprets EEG decoding as a similarity-based retrieval problem over a fixed semantic vocabulary.

\subsection{Optimization Strategies}
The semantic target space is highly sparse, with about 5 active concepts per sample out of 1210 possible tokens (99.6\% negatives). Under this extreme imbalance, standard optimization favors predicting the dominant negative class, leading to degenerate solutions that suppress most vocabulary items. To mitigate this, we employ complementary loss functions that encourage accurate concept selection while remaining robust to extreme class imbalance.

\textbf{Binary Cross-Entropy (BCE):}  
We frame the prediction of each concept $j \in \mathcal{V}$ as an independent binary classification task. To resolve selection difficulties caused by imbalance, we introduce a learned scaling parameter to the sigmoid function to improve separation between active and inactive concepts; the same adjustment is applied to our Focal Loss objective.

\textbf{Multi-Label Contrastive Loss:}  
We adapt the InfoNCE loss to a multi-label regime, treating all present concepts $\mathcal{P}$ as shared positive anchors:

\begin{equation}
\mathcal{L}_{\text{con}} =
-
\frac{1}{|\mathcal{P}|}
\sum_{j \in \mathcal{P}}
\log
\left(
\frac{\exp(\hat{y}_j / \tau)}
{\sum_{k=1}^{V} \exp(\hat{y}_k / \tau)}
\right)
\end{equation}

with temperature $\tau = 0.07$. This objective encourages the model to allocate probability mass to semantically relevant concepts while suppressing unrelated vocabulary items.

\textbf{Focal Loss:}  
To explicitly mitigate the overwhelming background of absent concepts, we apply Focal Loss with focusing parameter $\gamma = 2.0$ and weighting factor $\alpha_t = 0.25$. This loss emphasizes rare positive labels and difficult predictions, improving robustness in the extremely sparse label regime.

\subsection{Inference and LLM Integration}

During inference, we extract the top-$k$ ($k=15$) tokens from $\mathbf{\hat{y}}$ to form a Bag-of-Words representation $\mathcal{W}_{\text{top-}15}$. Selecting a small set of high-confidence concepts provides a stable semantic signal while filtering noisy predictions common in EEG decoding. This BoW is integrated with the primary object label ($o_{\text{pred}}$) and confidence score ($c_{\text{pred}}$) from the Stage-1 encoder into a structured zero-shot prompt. Instruction-tuned LLMs use these anchors as grounded priors to synthesize a syntactically fluent 8--20 word caption. This modular design ensures that raw neural data remain local while only discrete semantic cues interact with language models.

\section{Experiments}
We now describe the dataset, implementation details, and evaluation criteria used to evaluate \textsc{SENSE}.

\subsection{Dataset and Pre-processing} 
We utilize the \textit{CVPR2017} EEG-ImageNet benchmark, which contains synchronized EEG recordings from six subjects viewing 2,000 images. Following \citet{mishra-etal-2025-thought2text}, we use the expanded version pairing each stimulus with a natural language caption. Since each image is associated with multiple EEG trials across subjects, the dataset expands to a total of $9,940$ samples, partitioned into training ($7,959$), validation ($1,994$), and test ($1,987$) splits for fair comparison.

To construct the semantic supervision used to train the \textit{Similarity Refiner}, we extract nouns, verbs, and adjectives from captions in the training split using a POS-tagging and lemmatization pipeline. This produces a 1210-dimensional global vocabulary $\mathcal{V}$ and subject-agnostic N-hot target vectors $\mathbf{y} \in \{0,1\}^{1210}$. Restricting vocabulary construction to the training split prevents semantic leakage from the test set while enabling the model to learn generalized cross-subject mappings.


\subsection{Implementation Details}
The EEG encoder is pretrained using the ChannelNet architecture following \cite{mishra-etal-2025-thought2text} and remains frozen during training; only the \textit{Similarity Refiner} and associated projection components are optimized. The ChannelNet encoder contains $5,226,970$ parameters, while the refiner and other trainable components together comprise $1,052,161$ parameters. Training was conducted on a single NVIDIA A100 GPU using the AdamW optimizer ($lr=1\times10^{-4}$, weight decay $=1\times10^{-2}$) with a Cosine Annealing scheduler. Sigmoid-based variants (BCE and Focal Loss) converged within 50 epochs, whereas the Contrastive Multi-Label variant required 100 epochs to stabilize gradients. Owing to the lightweight design of the trainable module, overall training completes within a few minutes. To benchmark suitability for commodity hardware deployment, a full 50-epoch optimization run completes in $300.27$\,s ($\sim$6\,s/epoch) on a standard CPU with 8GB RAM, and local on-device inference requires $\sim$526\,ms/sample. For caption generation, we utilize API-based access to instruction-tuned LLMs: GPT-4o-mini and Gemini 2.5 Flash Lite, alongside \textsc{Qwen2.5-7B} and \textsc{LLaMa-3-8B} via their hosting on Together AI.

\begin{table*}[t]
    \centering
    \Description{A quantitative results table comparing four Similarity Refiner variants—Naive Baseline, Binary Cross Entropy, Contrastive Multi Label, and Focal Loss—across four LLM decoders: GPT-4o-mini, Gemini 2.5 Flash Lite, LLaMa-3-8B, and Qwen2.5-7B. The table reports ROUGE-1/2/L, BLEU-1/4, METEOR, BERT Score, and GPT-5-based Fluency and Adequacy scores. Each decoder is tested with and without the primary object label (WithObj vs. WithoutObj). A final section provides a performance comparison against the fine-tuned Thought2Text baseline. The data demonstrates that Focal Loss paired with Gemini 2.5 Flash Lite (WithObj) achieves the highest performance among the proposed SENSE configurations, matching or exceeding the fine-tuned Thought2Text baseline across most semantic and lexical overlap metrics.}
    \footnotesize
    \resizebox{\textwidth}{!}{
    \begin{tabular}{ll cc c cc c c | cc} 
         \toprule
         \textbf{Similarity Refiner} & \textbf{Decoder} & \multicolumn{2}{c}{\textbf{ROUGE-N (\%)}} & \textbf{ROUGE-L (\%)} & \multicolumn{2}{c}{\textbf{BLEU-N (\%)}} & \textbf{METEOR (\%)} & \textbf{BERT Score} & \multicolumn{2}{c}{\textbf{GPT-5}}\\ 
         Loss Function Variant & & \shortstack{N=1 \\ \scriptsize{$\pm$ 95\% CI}} & \shortstack{N=2 \\ \scriptsize{$\pm$ 95\% CI}} & \shortstack{ \\ \scriptsize{$\pm$ 95\% CI}} & \shortstack{N=1 \\ \scriptsize{$\pm$ 95\% CI}} & \shortstack{N=4 \\ \scriptsize{$\pm$ 95\% CI}} & \shortstack{ \\ \scriptsize{$\pm$ 95\% CI}} & & Flu. & Ade. \\ 
         \midrule

         \multirow{8}{*}{\textbf{Naive Baseline}} 
         & $\text{ChatGPT}_{\text{w}}$ & 25.8 [$\pm$0.5] & 4.8 [$\pm$0.3] & 23.8 [$\pm$0.5] & 21.6 [$\pm$0.5] & 3.8 [$\pm$0.2] & 20.2 [$\pm$0.5] & 0.888  & 4.59 & 1.22\\
         & $\text{ChatGPT}_{\text{wo}}$ & 22.7 [$\pm$0.4] & 3.1 [$\pm$0.2] & 20.7 [$\pm$0.4] & 20.0 [$\pm$0.4] & 3.1 [$\pm$0.1] & 17.1 [$\pm$0.4] & 0.877  & 4.58 & 1.28\\
         & $\text{Gemini}_{\text{w}}$ & 25.8 [$\pm$0.6] & 5.2 [$\pm$0.4] & 24.5 [$\pm$0.6] & 18.1 [$\pm$0.6] & 3.6 [$\pm$0.2] & 19.3 [$\pm$0.5] & 0.888  & 4.70 & 1.28\\
         & $\text{Gemini}_{\text{wo}}$ & 23.9 [$\pm$0.5] & 3.8 [$\pm$0.3] & 22.3 [$\pm$0.4] & 18.9 [$\pm$0.5] & 3.1 [$\pm$0.1] & 16.9 [$\pm$0.4] & 0.879  & 4.45 & 1.07\\
         & $\text{LLaMa}_{\text{w}}$ & 10.5 [$\pm$0.5] & 1.2 [$\pm$0.2] & 9.9 [$\pm$0.4] & 5.3 [$\pm$0.4] & 1.0 [$\pm$0.1] & 10.6 [$\pm$0.3] & 0.877  & 4.24 & 1.22\\
         & $\text{LLaMa}_{\text{wo}}$ & 13.7 [$\pm$0.4] & 1.8 [$\pm$0.2] & 12.4 [$\pm$0.4] & 10.8 [$\pm$0.4] & 1.9 [$\pm$0.1] & 13.0 [$\pm$0.3] & 0.866  & 3.84 & 1.04\\
         & $\text{Qwen}_{\text{w}}$ & 17.1 [$\pm$0.6] & 3.6 [$\pm$0.3] & 15.9 [$\pm$0.6] & 10.0 [$\pm$0.5] & 2.1 [$\pm$0.1] & 10.2 [$\pm$0.4] & 0.862  & 3.02 & 1.16\\
         & $\text{Qwen}_{\text{wo}}$ & 10.0 [$\pm$0.5] & 1.4 [$\pm$0.2] & 9.0 [$\pm$0.4] & 6.5 [$\pm$0.4] & 1.3 [$\pm$0.1] & 6.0 [$\pm$0.3] & 0.851  & 2.56 & 1.04\\
         \midrule

        \multirow{8}{*}{\textbf{Binary Cross Entropy}} 
         & $\text{ChatGPT}_{\text{w}}$ & 29.6 [$\pm$0.7] & 7.4 [$\pm$0.5] & 27.1 [$\pm$0.6] & 23.7 [$\pm$0.6] & 5.1 [$\pm$0.3] & 24.4 [$\pm$0.7] & \textit{\underline{0.897}}  & 4.71 & 1.38\\
         & $\text{ChatGPT}_{\text{wo}}$ & 29.4 [$\pm$0.6] & 6.6 [$\pm$0.4] & 26.2 [$\pm$0.5] & 24.2 [$\pm$0.6] & 4.5 [$\pm$0.2] & 23.5 [$\pm$0.6] & 0.893  & 4.58 & 1.28\\
         & $\text{Gemini}_{\text{w}}$ & \textit{30.8 [$\pm$0.7]} & 7.7 [$\pm$0.5] & 27.7 [$\pm$0.6] & 24.4 [$\pm$0.7] & 5.2 [$\pm$0.3] & 25.1 [$\pm$0.7] & 0.894   & 4.75 & \textit{\underline{1.40}}\\
         & $\text{Gemini}_{\text{wo}}$ & 29.1 [$\pm$0.6] & 6.2 [$\pm$0.4] & 25.9 [$\pm$0.6] & 23.4 [$\pm$0.6] & 4.4 [$\pm$0.2] & 23.1 [$\pm$0.6] & 0.889  & 4.72 & 1.31\\
         & $\text{LLaMa}_{\text{w}}$ & 12.4 [$\pm$0.6] & 1.8 [$\pm$0.2] & 11.3 [$\pm$0.5] & 6.6 [$\pm$0.4] & 1.3 [$\pm$0.1] & 12.1 [$\pm$0.4] & 0.880  & 4.54 & 1.30\\
         & $\text{LLaMa}_{\text{wo}}$ & 14.3 [$\pm$0.5] & 2.2 [$\pm$0.2] & 12.7 [$\pm$0.5] & 9.4 [$\pm$0.5] & 1.8 [$\pm$0.1] & 13.7 [$\pm$0.5] & 0.879  & 4.49 & 1.22\\
         & $\text{Qwen}_{\text{w}}$ & 20.1 [$\pm$0.7] & 5.4 [$\pm$0.4] & 18.1 [$\pm$0.6] & 12.7 [$\pm$0.6] & 3.1 [$\pm$0.2] & 13.9 [$\pm$0.6] & 0.875  & 3.27 & 1.33\\
         & $\text{Qwen}_{\text{wo}}$ & 18.7 [$\pm$0.6] & 4.0 [$\pm$0.3] & 16.3 [$\pm$0.5] & 12.8 [$\pm$0.5] & 2.6 [$\pm$0.2] & 14.0 [$\pm$0.5] & 0.869  & 3.20 & 1.25\\
         \midrule
         
         \multirow{8}{*}{\textbf{Contrastive Multi Label}} 
         & $\text{ChatGPT}_{\text{w}}$ & 29.2 [$\pm$0.6] & 7.7 [$\pm$0.5] & 27.0 [$\pm$0.6] & 23.8 [$\pm$0.6] & 5.3 [$\pm$0.3] & 24.1 [$\pm$0.7] & 0.895  & 4.70 & 1.38\\
         & $\text{ChatGPT}_{\text{wo}}$ & 28.3 [$\pm$0.6] & 6.2 [$\pm$0.4] & 25.5 [$\pm$0.5] & 24.1 [$\pm$0.6] & 4.4 [$\pm$0.2] & 22.8 [$\pm$0.6] & 0.891  & 4.64 & 1.29\\
         & $\text{Gemini}_{\text{w}}$ & 30.2 [$\pm$0.7] & 7.7 [$\pm$0.5] & 27.3 [$\pm$0.6] & 24.4 [$\pm$0.7] & 5.2 [$\pm$0.3] & 24.9 [$\pm$0.7] & 0.894  & \textit{\underline{4.77}} & \textit{\underline{1.40}}\\
         & $\text{Gemini}_{\text{wo}}$ & 27.7 [$\pm$0.6] & 6.2 [$\pm$0.4] & 25.0 [$\pm$0.5] & 23.0 [$\pm$0.6] & 4.3 [$\pm$0.2] & 22.8 [$\pm$0.6] & 0.888  & 4.71 & 1.30\\
         & $\text{LLaMa}_{\text{w}}$ & 12.3 [$\pm$0.5] & 1.8 [$\pm$0.2] & 11.4 [$\pm$0.5] & 6.8 [$\pm$0.4] & 1.3 [$\pm$0.1] & 12.2 [$\pm$0.4] & 0.880  & 4.50 & 1.31\\
         & $\text{LLaMa}_{\text{wo}}$ & 15.1 [$\pm$0.5] & 2.5 [$\pm$0.2] & 13.7 [$\pm$0.5] & 10.1 [$\pm$0.5] & 1.9 [$\pm$0.1] & 14.2 [$\pm$0.5] & 0.879  & 4.47 & 1.24\\
         & $\text{Qwen}_{\text{w}}$ & 20.3 [$\pm$0.7] & 5.4 [$\pm$0.4] & 18.6 [$\pm$0.6] & 12.8 [$\pm$0.6] & 3.0 [$\pm$0.2] & 13.8 [$\pm$0.6] & 0.873  & 3.27 & 1.34\\
         & $\text{Qwen}_{\text{wo}}$ & 17.5 [$\pm$0.6] & 3.9 [$\pm$0.3] & 15.6 [$\pm$0.5] & 12.0 [$\pm$0.5] & 2.5 [$\pm$0.2] & 12.9 [$\pm$0.5] & 0.869  & 3.23 & 1.25\\
         \midrule

         \multirow{8}{*}{\textbf{Focal}} 
         & $\text{ChatGPT}_{\text{w}}$ & 30.6 [$\pm$0.7] & \textit{8.2 [$\pm$0.5]} & \textit{28.2 [$\pm$0.6]} & 24.9 [$\pm$0.6] & \textbf{\underline{5.6 [$\pm$0.3]}} & 25.3 [$\pm$0.7] & \textbf{0.898}  & 4.75 & \textit{\underline{1.40}}\\
         & $\text{ChatGPT}_{\text{wo}}$ & 30.3 [$\pm$0.6] & 7.3 [$\pm$0.4] & 27.1 [$\pm$0.6] & 25.2 [$\pm$0.6] & 4.7 [$\pm$0.2] & 24.5 [$\pm$0.6] & 0.895  & 4.60 & 1.27\\
         & $\text{Gemini}_{\text{w}}$ & \textbf{31.5 [$\pm$0.7]} & \textbf{8.5 [$\pm$0.5]} & \textbf{28.7 [$\pm$0.7]} & \textit{25.2 [$\pm$0.7]} & \textbf{\underline{5.6 [$\pm$0.3]}} & \textit{26.1 [$\pm$0.8]} & \textit{\underline{0.897}}  & \textit{\underline{4.77}} &  \textit{\underline{1.40}} \\
         & $\text{Gemini}_{\text{wo}}$ & 30.5 [$\pm$0.6] & 7.6 [$\pm$0.4] & 27.1 [$\pm$0.6] & 24.7 [$\pm$0.6] & 4.9 [$\pm$0.3] & 24.4 [$\pm$0.7] & 0.893  & 4.67 & 1.30\\
         & $\text{LLaMa}_{\text{w}}$ & 12.6 [$\pm$0.6] & 2.2 [$\pm$0.3] & 11.6 [$\pm$0.5] & 6.8 [$\pm$0.5] & 1.5 [$\pm$0.2] & 12.5 [$\pm$0.5] & 0.881  & 4.56 & 1.33 \\
         & $\text{LLaMa}_{\text{wo}}$ & 15.9 [$\pm$0.6] & 2.7 [$\pm$0.3] & 14.3 [$\pm$0.5] & 10.4 [$\pm$0.5] & 2.1 [$\pm$0.1] & 14.8 [$\pm$0.5] & 0.882  & 4.46 & 1.24\\
         & $\text{Qwen}_{\text{w}}$ & 21.8 [$\pm$0.7] & 5.9 [$\pm$0.4] & 19.9 [$\pm$0.6] & 13.8 [$\pm$0.6] & 3.2 [$\pm$0.2] & 15.0 [$\pm$0.6] & 0.876  & 3.47 & 1.33\\
         & $\text{Qwen}_{\text{wo}}$ & 20.2 [$\pm$0.6] & 4.5 [$\pm$0.3] & 17.7 [$\pm$0.5] & 14.2 [$\pm$0.5] & 2.9 [$\pm$0.2] & 15.4 [$\pm$0.5] & 0.874  & 3.41 & 1.24\\
         \midrule

         \multirow{2}{*}{\textbf{Thought2Text}} 
         & $\mathrm{LLaMa_{ALL}}$ & 30.0 & 8.1 & 26.6 & \textbf{25.5} & \textit{5.5} & \textbf{26.3} & 0.89 & \textbf{4.82} & \textbf{1.58}\\
         & $\mathrm{Qwen_{ALL}}$ & 26.4 & 4.6 & 22.8 & 22.7 & 3.7 & 21.1 & 0.88 & 4.75 & 1.28\\
         \bottomrule
    \end{tabular}}
    \caption{Average evaluation metrics and GPT-5 assessment (\textit{Flu.}: fluency, \textit{Ade.}: adequacy) for text generated from Bag-of-Words (BoW) across four LLMs: GPT-4o-mini, Gemini 2.5 Flash Lite, LLaMa-3-8B, and Qwen2.5-7B. Decoders are appended with subscripts indicating configurations evaluated with visual grounding context ($\text{w}$ for \textsc{WithObj}), solely from neural BoW ($\text{wo}$ for \textsc{WithoutObj}), or under fully fine-tuned baselines ($\text{all}$). For ROUGE-1/2/L, BLEU-1/4, METEOR and BERT Score, metrics report the Mean $\pm$ the 95\% Bootstrap Margin of Error calculated over 10,000 resamples; for BERT Score, the 95\% Bootstrap Margin of Error is uniformly $\pm 0.001$ across all settings. For all metrics, higher values indicate superior performance. The best results within evaluation domains are \textbf{bolded} and second-best are \textit{italicized}; matching ties are \underline{underlined}.}
    \label{tab:results}
\end{table*}

\subsection{LLM Prompting Strategy}
We employ a zero-shot prompting strategy that provides the LLM with: (1) the primary object label and confidence score $c_{\text{pred}}$ from the Stage-1 encoder, and (2) the top-15 BoW tokens with their corresponding cosine similarity scores. The model is instructed to treat the object label as a grounding anchor and use the similarity scores to weigh the relative importance of BoW concepts while generating a cohesive caption of 8--20 words. To isolate the contribution of the object anchor, we additionally conduct an ablation (\textit{WithoutObj}) in which all instructions and inputs related to $o_{\text{pred}}$ are removed, forcing the LLM to reconstruct captions solely from the EEG-derived BoW. All LLMs are used in their standard configuration without invoking additional reasoning modes or specialized capabilities, and generation uses a fixed $0.2$ temperature across models for grounded reproducibility; other decoding parameters (e.g., top-$p$) remain at default values. Due to space limitations, example prompts are provided in the appendix.


\subsection{Evaluation Metrics}
Following standard practice in natural language generation, we report BLEU-1/4 \cite{papineni2002bleu} and ROUGE-1/2/L \cite{lin2004rouge} to measure lexical overlap between generated captions and references. To better capture semantic alignment and contextual similarity, we additionally compute METEOR \cite{banerjee2005meteor} and BERTScore \cite{zhang2019bertscore}. In addition to automated metrics, we perform LLM-based evaluation, using GPT-5 to assess generation quality along two dimensions: \textit{fluency}, which measures grammatical correctness and readability, and \textit{adequacy}, which measures how accurately the generated caption conveys the meaning of the reference. Each dimension is rated on a 1--5 scale, where 5 indicates the highest quality. Scores reported in Table~\ref{tab:results} are averaged across all samples in the test split. We explicitly acknowledge that utilizing a proprietary, updateable API-based model as an automated judge introduces inherent evaluation instability and reproducibility limits over time. While generation parameters were held strict (e.g., locking temperature at 0.2), backend updates or version rolling by the provider could subtly alter scoring behavior compared to static human annotation or open-weight models.

\section{Results}

\begin{figure*}[t]
\Description{A composite figure consisting of six radar (spider) charts arranged in two rows and three columns, corresponding to Subjects 1 through 6. Each chart plots BLEU-1 scores across four axes representing optimization strategies: Focal Loss (top), Contrastive Multi-Label (right), BCE Loss (bottom), and Naive Baseline (left). The radial axes are scaled from 0.1 to 0.4. Four colored polygons represent different LLM decoders: blue for Gemini, red for ChatGPT, orange for LLaMa-3-8B, and green for Qwen2.5-7B. In every subject chart, the blue and red polygons are the outermost, peaking near 0.4 for Focal Loss and Contrastive ML, indicating superior performance. The green and orange polygons are consistently nested inside, reflecting lower scores. The geometric shape of the performance polygons is highly consistent across all six subjects, demonstrating stable cross-subject generalization. A shared legend at the bottom identifies the four models by color, and the 4 training strategies.}
    \centering
    
    \begin{subfigure}{0.31\textwidth}
        \centering
        \Description{Radar chart for Subject 1.}
        \includegraphics[width=\linewidth]{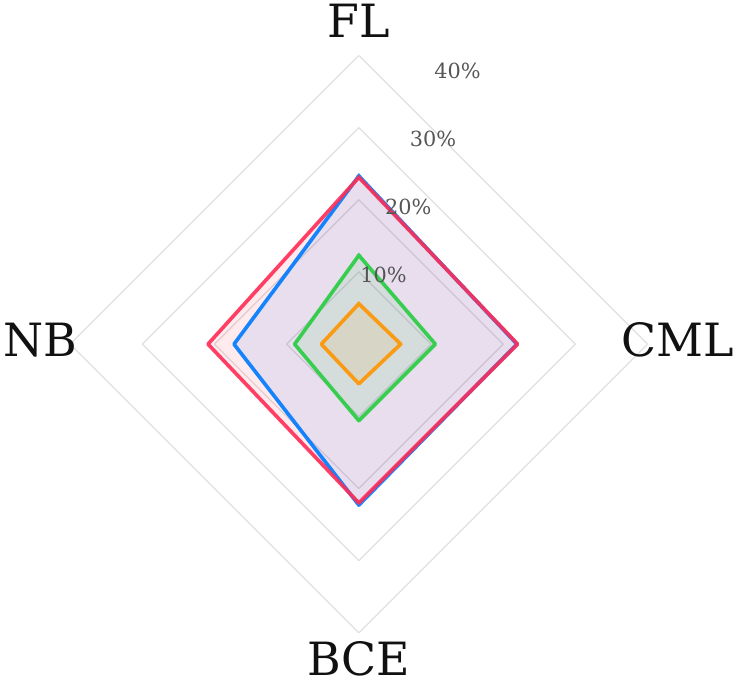}
        \vspace{-8pt} 
        \caption{Subject 1} 
        \label{fig:subj1}
    \end{subfigure}
    \hfill 
    \begin{subfigure}{0.31\textwidth}
        \centering
        \includegraphics[width=\linewidth]{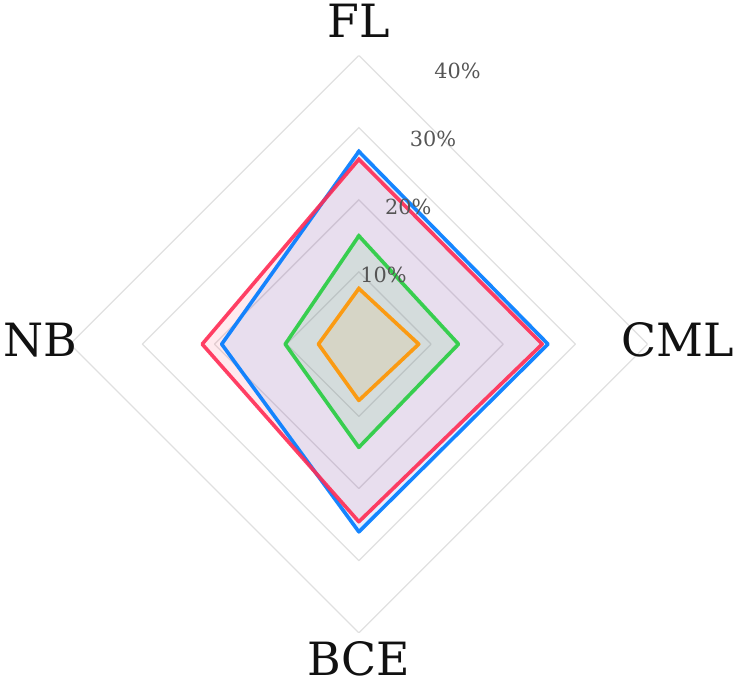}
        \vspace{-8pt}
        \caption{Subject 2}
        \label{fig:subj2}
    \end{subfigure}
    \hfill
    \begin{subfigure}{0.31\textwidth}
        \centering
        \includegraphics[width=\linewidth]{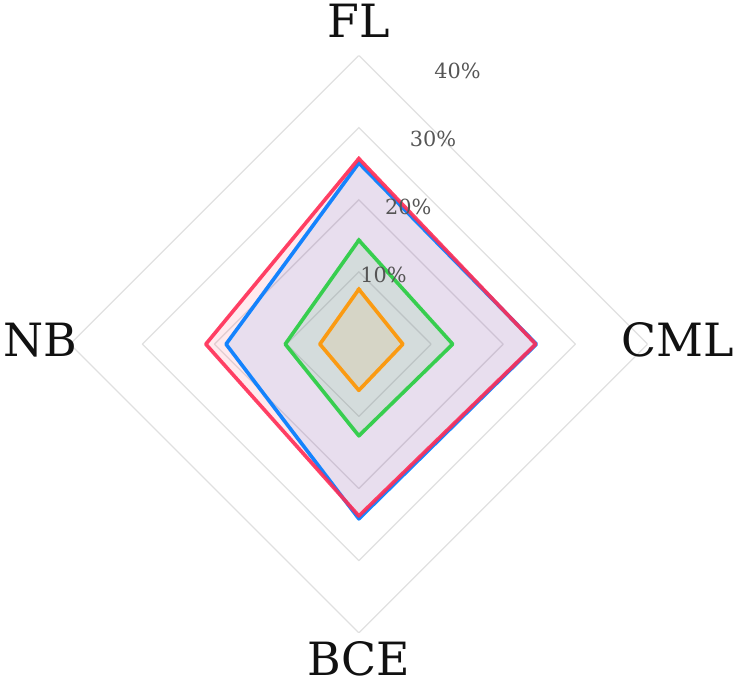}
        \vspace{-8pt}
        \caption{Subject 3}
        \label{fig:subj3}
    \end{subfigure}

    \vspace{2pt} 

    \begin{subfigure}{0.31\textwidth}
        \centering
        \includegraphics[width=\linewidth]{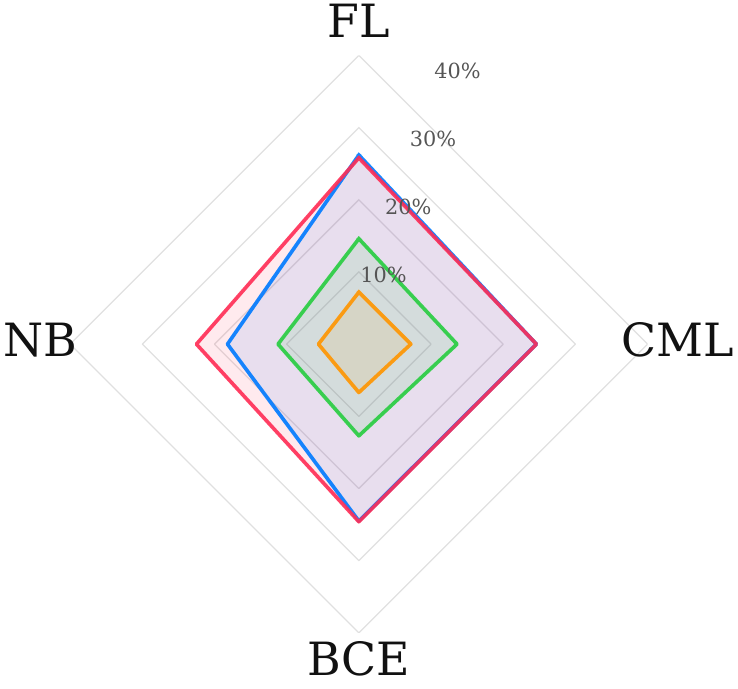}
        \vspace{-8pt}
        \caption{Subject 4}
        \label{fig:subj4}
    \end{subfigure}
    \hfill
    \begin{subfigure}{0.31\textwidth}
        \centering
        \includegraphics[width=\linewidth]{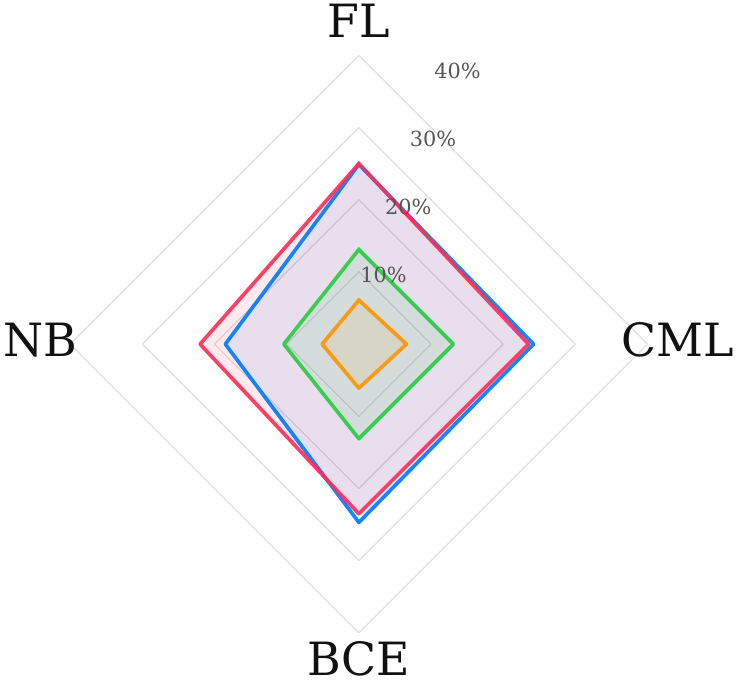}
        \vspace{-8pt}
        \caption{Subject 5}
        \label{fig:subj5}
    \end{subfigure}
    \hfill
    \begin{subfigure}{0.31\textwidth}
        \centering
        \includegraphics[width=\linewidth]{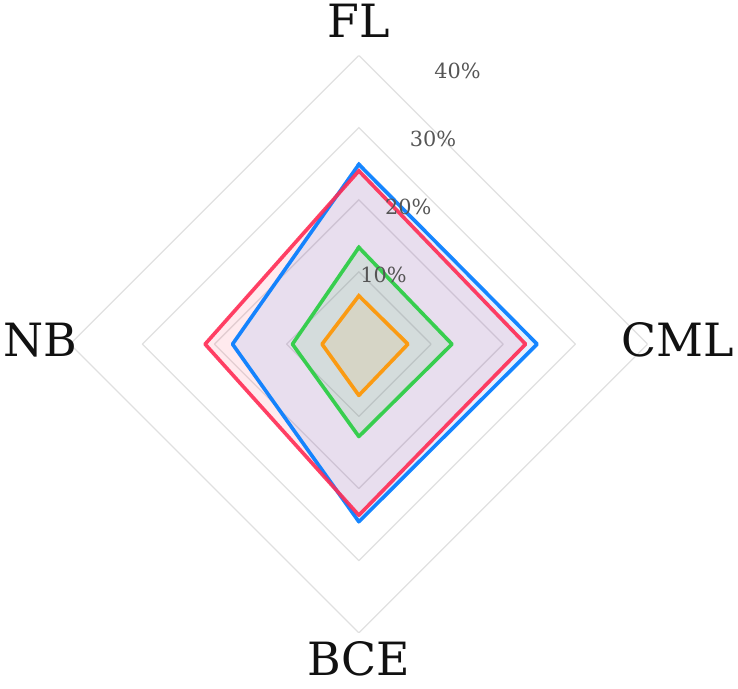}
        \vspace{-8pt}
        \caption{Subject 6}
        \label{fig:subj6}
    \end{subfigure}

    \vspace{-4pt} 
    
    {\centering
    \includegraphics[width=0.7\textwidth]{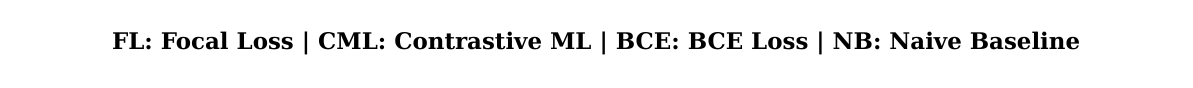} \par}
    
    \vspace{-4pt} 

    {\centering
    \includegraphics[width=0.7\textwidth]{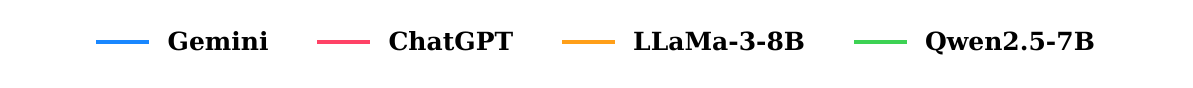} \par}
    
    \vspace{-4pt} 
    \caption{Subject-wise BLEU-1 (\%) performance across optimization strategies and LLM decoders. The radar charts demonstrate that premier closed-source models (Gemini and ChatGPT) consistently form the outer performance boundaries across all six subjects, with ChatGPT exhibiting a slight advantage in the Naive Baseline task. Furthermore, the topological consistency of the polygons indicates that our framework generalizes effectively across diverse neural patterns without requiring per-subject fine-tuning.}
    \label{fig:all_spider_plots}
\end{figure*}

Table \ref{tab:results} summarizes performance across training loss regimes and LLM decoders. Focal Loss performs best, while the Naive Baseline performs poorly. Gemini 2.5 Flash Lite consistently outperforms other decoders, except in the Naive Baseline where GPT-4o-mini has a slight edge. 


The results also highlight architectural differences: closed-source models effectively leverage structured Bag-of-Words prompts to synthesize fluent text, whereas un-finetuned open-source models like \textsc{LLaMa-3-8B} and \textsc{Qwen2.5-7B} struggle to integrate discrete semantic anchors. This performance gap stems primarily from the fact that smaller, open-weight models have strict in-context alignment boundaries; without explicit task-specific fine-tuning on noisy inputs, they struggle to map abstract token bags to rigid syntactic structures, often prioritizing their internal linguistic priors over the sparse conditioning anchors. Notably, \textsc{LLaMa-3-8B} uniquely improves in the \textsc{WithoutObj} setting, suggesting greater reliance on distributed semantic cues from the BoW. Overall, these models score near zero on exact-match metrics like BLEU-4 but maintain moderately high BERT scores, indicating an ability to capture general semantics without precise syntactic reconstruction.

Despite using a lightweight EEG-to-keyword module and prompt-based generation, our approach achieves performance comparable to, and sometimes exceeding, fine-tuned systems such as \textsc{Thought2Text} (e.g., 31.5 vs. 30.0 ROUGE-1 with $\mathrm{LLaMa\textrm{-}3\textrm{-}8B_{ALL}}$). Because the neural decoding stage operates locally and exposes only abstract semantic tokens to the LLM, raw EEG signals remain on-device, supporting a privacy-preserving design.

\subsection{Subject-Wise Performance Analysis}
A primary challenge in non-invasive BCI decoding is the inherent inter-subject variability of EEG signals. To evaluate the \textit{Similarity Refiner}'s robustness, a subject-wise performance analysis of BLEU-1 scores was conducted across all six participants (Figure \ref{fig:all_spider_plots}). The resulting radar charts reveal remarkable topological consistency, indicating that the hierarchy of optimization strategies holds true regardless of individual neural patterns. Across all subjects, Focal loss achieves the highest performance, while the Naive Baseline and BCE loss remain contracted. Additionally, premier closed-source models (Gemini 2.5 Flash Lite and GPT-4o-mini) consistently form the outer performance boundaries. Ultimately, the framework demonstrates an impressive ability to generalize a single, subject-agnostic projection matrix across diverse human participants without requiring individualized, per-subject calibration.

\newcommand{\unsmush}[1]{\makecell{\vspace{-5pt} \\ #1 \\ \vspace{-5pt}}}
\begin{table*}
\Description{A qualitative comparison table with 8 rows. Each row contains an image, its reference object, predicted object, reference description, a Thought2Text baseline caption, a Sample Bag-of-Words, and the SENSE Description. Row 1-4 show successful retrievals (mushroom, piano, camping, pumpkin) where SENSE captions are highlighted in green and match the semantic content of the images. Row 5-8 show error cases (flower misidentified as mushroom, coffee mug, guitar misidentified as train, and piano misidentified as ball) highlighted in red. In the error cases, the SENSE description remains fluent but reflects the incorrect object label provided by the refiner, illustrating the system's reliance on accurate semantic cues.}
\centering
\small
\setlength{\tabcolsep}{3pt}

\begin{tabular}{c c c c c c c c}
\hline
\textbf{ID} & \textbf{Image} & \textbf{Ref. Obj} & \textbf{Pred. Obj} & \textbf{Reference Description} & \textbf{\textsc{Thought2Text}} & \textbf{Sample BoW} & \textbf{\textsc{SENSE} Description} \\ \hline

1 & 
\unsmush{\includegraphics[width=0.10\textwidth]{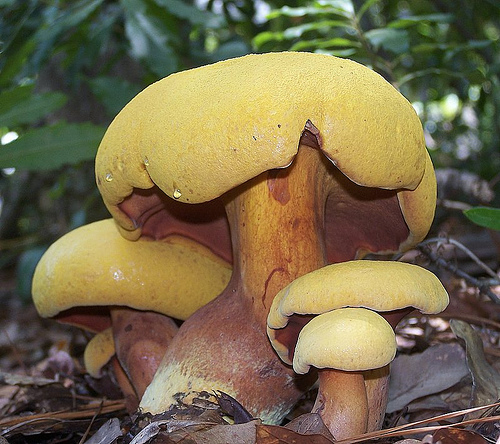}} & 
\makecell{mushroom} & 
\makecell{mushroom} & 
\makecell{\begin{tabular}{p{0.14\textwidth}}\centering A large yellow mushroom with a brown stem and a brown cap, surrounded by green foliage.\end{tabular}} & 
\makecell{\begin{tabular}{p{0.14\textwidth}}\centering A group of mushrooms growing on a log.\end{tabular}} & 
\makecell{\begin{tabular}{p{0.08\textwidth}}\centering yellow, mushroom, white, daisy, flower\end{tabular}} & 
\makecell{\begin{tabular}{p{0.16\textwidth}}\centering \textcolor{ACMGreen}{A yellow mushroom growing in the green grass.}\end{tabular}} \\ \hline

2 & 
\unsmush{\includegraphics[width=0.10\textwidth]{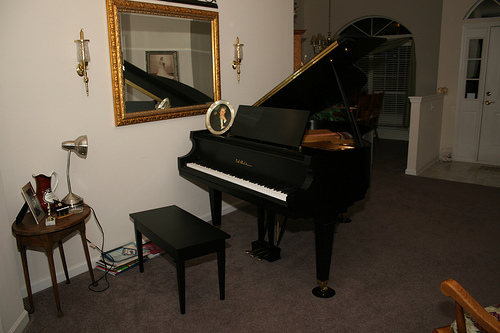}} & 
\makecell{piano} & 
\makecell{piano} & 
\makecell{\begin{tabular}{p{0.14\textwidth}}\centering A black grand piano in a living room.\end{tabular}} & 
\makecell{\begin{tabular}{p{0.14\textwidth}}\centering A grand piano with a stool in front of it.\end{tabular}} & 
\makecell{\begin{tabular}{p{0.08\textwidth}}\centering piano, black, room, grand, floor\end{tabular}} & 
\makecell{\begin{tabular}{p{0.16\textwidth}}\centering \textcolor{ACMGreen}{A grand piano on a wooden floor.}\end{tabular}} \\ \hline

3 & 
\unsmush{\includegraphics[width=0.10\textwidth]{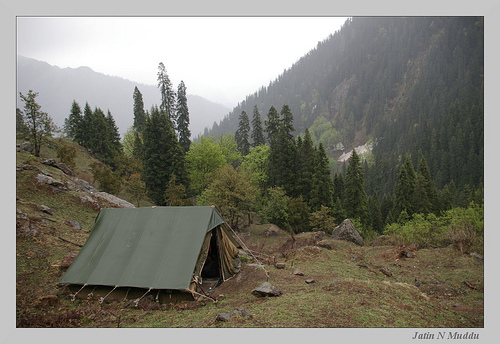}} & 
\makecell{camping} & 
\makecell{camping} & 
\makecell{\begin{tabular}{p{0.14\textwidth}}\centering A tent in a mountainous area with trees and fog.\end{tabular}} & 
\makecell{\begin{tabular}{p{0.14\textwidth}}\centering A tent set up in a forest with a campfire nearby. \end{tabular}} & 
\makecell{\begin{tabular}{p{0.08\textwidth}}\centering tent, field, grassy, lawn\end{tabular}} & 
\makecell{\begin{tabular}{p{0.16\textwidth}}\centering \textcolor{ACMGreen}{A tent is set up in a grassy field for camping.}\end{tabular}} \\ \hline

4 & 
\unsmush{\includegraphics[width=0.10\textwidth]{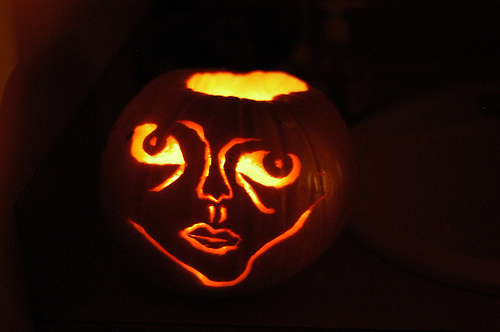}} & 
\makecell{pumpkin} & 
\makecell{pumpkin} & 
\makecell{\begin{tabular}{p{0.14\textwidth}}\centering A carved pumpkin with a face and eyes, sitting on a table.\end{tabular}} & 
\makecell{\begin{tabular}{p{0.14\textwidth}}\centering A carved pumpkin with a spooky face on it.\end{tabular}} & 
\makecell{\begin{tabular}{p{0.08\textwidth}}\centering carved, pumpkin, face, halloween\end{tabular}} & 
\makecell{\begin{tabular}{p{0.16\textwidth}}\centering \textcolor{ACMGreen}{A carved pumpkin with a spooky face sits on the porch.}\end{tabular}} \\ \hline

5 & 
\unsmush{\includegraphics[width=0.10\textwidth]{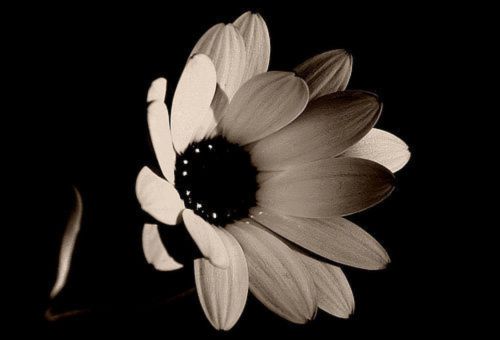}} & 
\makecell{flower} & 
\makecell{mushroom} & 
\makecell{\begin{tabular}{p{0.14\textwidth}}\centering A black and white photograph of a single daisy with a white center and a dark brown center.\end{tabular}} & 
\makecell{\begin{tabular}{p{0.14\textwidth}}\centering A group of mushrooms growing on a log.\end{tabular}} & 
\makecell{\begin{tabular}{p{0.08\textwidth}}\centering mushroom, grow, white, forest, red\end{tabular}} & 
\makecell{\begin{tabular}{p{0.16\textwidth}}\centering \textcolor{ACMRed}{A white mushroom grows on the forest ground.}\end{tabular}} \\ \hline

6 & 
\unsmush{\includegraphics[width=0.10\textwidth]{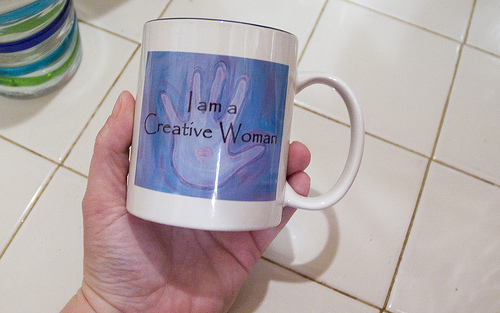}} & 
\makecell{coffee mug} & 
\makecell{coffee mug} & 
\makecell{\begin{tabular}{p{0.14\textwidth}}\centering A hand holding a mug with a blue background and a handprint design. \end{tabular}} & 
\makecell{\begin{tabular}{p{0.14\textwidth}}\centering A person holding a coffee mug with a A person holding a coffee mug with the words ``World's Best Dad'' written on it.\end{tabular}} & 
\makecell{\begin{tabular}{p{0.08\textwidth}}\centering mug, coffee, white, black, cup\end{tabular}} & 
\makecell{\begin{tabular}{p{0.16\textwidth}}\centering \textcolor{ACMRed}{A white coffee mug with a black logo on a table.}\end{tabular}} \\ \hline

7 & 
\unsmush{\includegraphics[width=0.10\textwidth]{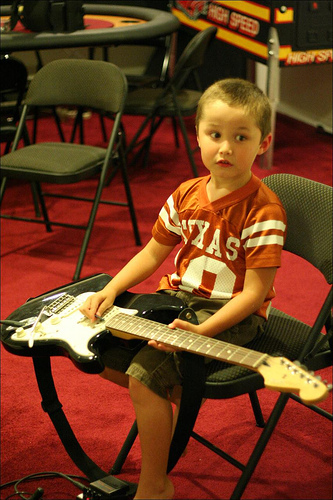}} & 
\makecell{guitar} & 
\makecell{train} & 
\makecell{\begin{tabular}{p{0.14\textwidth}}\centering A young boy sitting on a chair playing a guitar.\end{tabular}} & 
\makecell{\begin{tabular}{p{0.14\textwidth}}\centering A man is holding a guitar in front of a microphone.\end{tabular}} & 
\makecell{\begin{tabular}{p{0.08\textwidth}}\centering train, track, black, blue, white\end{tabular}} & 
\makecell{\begin{tabular}{p{0.16\textwidth}}\centering \textcolor{ACMRed}{A black and white truck on a platform.}\end{tabular}} \\ \hline

8 & 
\unsmush{\includegraphics[width=0.10\textwidth]{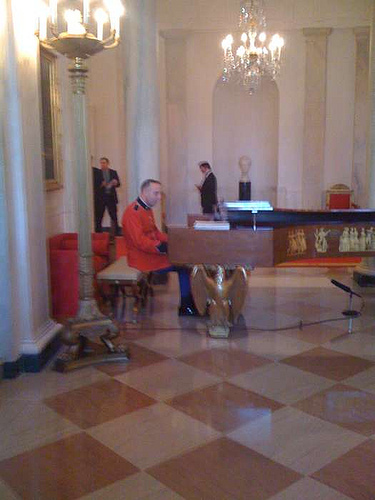}} & 
\makecell{piano} & 
\makecell{ball} & 
\makecell{\begin{tabular}{p{0.14\textwidth}}\centering A man in a red coat and black pants is playing a piano in a room with a chandelier.\end{tabular}} & 
\makecell{\begin{tabular}{p{0.14\textwidth}}\centering A man is playing the piano in a dimly lit room.\end{tabular}} & 
\makecell{\begin{tabular}{p{0.08\textwidth}}\centering piano, golf, ball, black, room\end{tabular}} & 
\makecell{\begin{tabular}{p{0.16\textwidth}}\centering \textcolor{ACMRed}{A black and white billiard table with a green felt.}\end{tabular}} \\ \hline

\end{tabular}

\caption{Qualitative comparison of generated descriptions from \textsc{SENSE} and the \textsc{Thought2Text} baseline. The Sample BoW column shows a representative subset of the Bag-of-Words provided to the LLM during caption generation. Samples 7 and 8 illustrate ablation results where no object label was supplied to the model. (\textit{SENSE captions generated by Gemini 2.5 Flash Lite}.)}
\label{tab:qualitative}
\end{table*}

\subsection{Qualitative Generation Analysis}
Table \ref{tab:qualitative} qualitatively compares captions generated by SENSE to the fine-tuned \textsc{Thought2Text} baseline. SENSE effectively leverages the extracted Bag-of-Words (BoW) and predicted object labels to synthesize fluent descriptions, frequently capturing richer peripheral details that rival or exceed the baseline's output. However, this modularity introduces specific generative vulnerabilities inherent to instruction-tuned LLMs, most notably semantic drift and hallucinated text.

Semantic drift occurs when the LLM over-extrapolates from the provided semantic anchors, weaving the discrete tokens into a cohesive but contextually inaccurate narrative. If the retrieved BoW contains semantically distant tokens due to noise in the retrieval stage, the LLM may construct a plausible but entirely fabricated scene composition to forcefully reconcile these disparate cues.

Additionally, the model occasionally hallucinates granular visual details (e.g., distinct colors, textures, or spatial relationships) not strictly grounded in the discrete semantic cues. For instance, incorrect Stage-1 object predictions (e.g., misidentifying a mushroom as a flower) heavily bias the LLM's generative prior. This bias not only skews the primary subject, but frequently cascades into hallucinated peripheral features assumed to co-occur with the misidentified object. This highlights a critical trade-off in modular BCI decoding: the LLM's powerful zero-shot capabilities drive SENSE's fluency, but they simultaneously risk overriding the authentic neural signal if the grounding anchors are flawed.

Nonetheless, ablation studies (Samples 7 and 8) demonstrate that even without the primary object anchor, the LLM still captures the general scene composition using only the BoW, despite occasionally struggling to pinpoint the exact primary subject.

\section{Conclusion and Future Work}

\label{sec:concl}
In this work, we introduced \textsc{SENSE}, a lightweight framework for EEG-to-text decoding that eliminates the need for fine-tuning large language models by decomposing the task into on-device semantic retrieval and prompt-based language generation. \textsc{SENSE} aligns continuous neural signals with a CLIP-grounded vocabulary to extract discrete semantic cues that condition an off-the-shelf LLM. A key finding of this work is that such a simple and modular approach can achieve performance comparable to, and in several cases exceeding, fully fine-tuned systems such as \textsc{Thought2Text}. This highlights that explicit semantic retrieval from EEG signals can effectively guide language generation without per-subject calibration or the heavy training pipelines required by prior approaches. The EEG-to-keyword module contains only $\sim$6M parameters and runs entirely on-device, ensuring that raw neural signals remain local while only abstract semantic cues interact with language models, reducing exposure of sensitive brain data and enabling a practical privacy-aware architecture for scalable BCI deployments. Future work will explore improving the utilization of the extracted BoW by replacing static top-$k$ retrieval with dynamic confidence thresholding, incorporating syntactic priors during concept selection, expanding the semantic vocabulary, strengthening cross-modal alignment with visual context, and enabling real-time multi-sensor fusion for more expressive brain-driven language interfaces.

\section{Safe and Responsible Innovation Statement}
\textsc{SENSE} is a lightweight, privacy-preserving framework for decoding EEG signals into text, intended to advance assistive communication for individuals with neurological disorders. By localizing neural feature extraction and semantic retrieval to the user's device, the architecture ensures that raw, high-dimensional brain signals never leave the local environment, mitigating significant privacy risks associated with transmitting sensitive neural data to external services. To address the risk of "ungrounded generation," \textsc{SENSE} enforces semantic grounding via an interpretable Bag-of-Words interface. Furthermore, our subject-agnostic projection matrix generalizes effectively across diverse neural patterns without individualized calibration, promoting equitable access and reducing potential bias. We remain committed to end-to-end data autonomy and user safety.

\subsection{Formal Threat Model and Leakage Analysis}
To validate our privacy guarantees, we formalize a threat model assuming a secure local trust boundary and an untrusted or semi-trusted cloud environment. The primary threat vector involves an adversary intercepting the transmitted top-$15$ keywords or a compromised cloud provider attempting to infer user attributes from the payload. We evaluate two leakage profiles:

\begin{enumerate}
    \item \textit{Neural State Reconstruction}: Raw EEG signals contain granular biometric signatures revealing identity or cognitive conditions. By mapping continuous signals into a local textual concept space, the physical properties of the neural data are fundamentally destroyed, preventing an adversary from reconstructing raw brain activity waveforms.
    \item \textit{Affective and Cognitive Profiling}: Token-based transmission introduces risks of leaking user mental states. However, \textsc{SENSE} mitigates this via vocabulary constraint. Because our global vocabulary is explicitly bounded by objective ImageNet annotations, retrieval targets are strictly limited to non-affective descriptions of external visual stimuli. Because information-dense neural embeddings remain local and the transmitted top-$k$ anchors lack emotional or subjective categories, reverse-engineering a user's cognitive state from the sparse visual payload is rendered impossible.
\end{enumerate}

\section*{Generative AI Disclosure}
Large Language Models (including Gemini and ChatGPT) were used to assist in \LaTeX\ formatting, ensuring sentence conciseness, and aiding in error debugging during the preparation of this work. All authors reviewed and are responsible for the final content.

\bibliographystyle{ACM-Reference-Format}
\clearpage
\bibliography{acmart}

@inproceedings{mishra-etal-2025-thought2text,
    title = "{T}hought2{T}ext: Text Generation from {EEG} Signal using Large Language Models ({LLM}s)",
    author = "Mishra, Abhijit  and
      Shukla, Shreya  and
      Torres, Jose  and
      Gwizdka, Jacek  and
      Roychowdhury, Shounak",
    editor = "Chiruzzo, Luis  and
      Ritter, Alan  and
      Wang, Lu",
    booktitle = "Findings of the Association for Computational Linguistics: NAACL 2025",
    month = apr,
    year = "2025",
    address = "Albuquerque, New Mexico",
    publisher = "Association for Computational Linguistics",
    url = "https://aclanthology.org/2025.findings-naacl.207/",
    doi = "10.18653/v1/2025.findings-naacl.207",
    pages = "3747--3759",
    ISBN = "979-8-89176-195-7"
}

@inproceedings{tao2025see,
  title={See: Semantically aligned eeg-to-text translation},
  author={Tao, Yitian and Liang, Yan and Wang, Luoyu and Li, Yongqing and Yang, Qing and Zhang, Han},
  booktitle={ICASSP 2025-2025 IEEE International Conference on Acoustics, Speech and Signal Processing (ICASSP)},
  pages={1--5},
  year={2025},
  organization={IEEE}
}

@article{masry2025ets,
  title={ETS: Open Vocabulary Electroencephalography-To-Text Decoding and Sentiment Classification},
  author={Masry, Mohamed and Amen, Mohamed and Elzyat, Mohamed and Hamed, Mohamed and Magdy, Norhan and Khaled, Maram},
  journal={arXiv preprint arXiv:2506.14783},
  year={2025}
}

@article{levy2025brain,
  title={Brain-to-text decoding: A non-invasive approach via typing},
  author={L{\'e}vy, Jarod and Zhang, Mingfang and Pinet, Svetlana and Rapin, J{\'e}r{\'e}my and Banville, Hubert and d'Ascoli, St{\'e}phane and King, Jean-R{\'e}mi},
  journal={arXiv preprint arXiv:2502.17480},
  year={2025}
}

@inproceedings{papineni2002bleu,
  title={Bleu: a method for automatic evaluation of machine translation},
  author={Papineni, Kishore and Roukos, Salim and Ward, Todd and Zhu, Wei-Jing},
  booktitle={Proceedings of the 40th annual meeting of the Association for Computational Linguistics},
  pages={311--318},
  year={2002}
}

@inproceedings{lin2004rouge,
  title={Rouge: A package for automatic evaluation of summaries},
  author={Lin, Chin-Yew},
  booktitle={Text summarization branches out},
  pages={74--81},
  year={2004}
}

@inproceedings{banerjee2005meteor,
    title = "{METEOR}: An Automatic Metric for {MT} Evaluation with Improved Correlation with Human Judgments",
    author = "Banerjee, Satanjeev  and
      Lavie, Alon",
    editor = "Goldstein, Jade  and
      Lavie, Alon  and
      Lin, Chin-Yew  and
      Voss, Clare",
    booktitle = "Proceedings of the {ACL} Workshop on Intrinsic and Extrinsic Evaluation Measures for Machine Translation and/or Summarization",
    month = jun,
    year = "2005",
    address = "Ann Arbor, Michigan",
    publisher = "Association for Computational Linguistics",
    url = "https://aclanthology.org/W05-0909/",
    pages = "65--72"
}

@article{zhang2019bertscore,
  title={Bertscore: Evaluating text generation with bert},
  author={Zhang, Tianyi and Kishore, Varsha and Wu, Felix and Weinberger, Kilian Q and Artzi, Yoav},
  journal={arXiv preprint arXiv:1904.09675},
  year={2019}
}

@inproceedings{spampinato2017deep,
  title={Deep learning human mind for automated visual classification},
  author={Spampinato, Concetto and Palazzo, Simone and Kavasidis, Isaak and Giordano, Daniela and Souly, Nasim and Shah, Mubarak},
  booktitle={Proceedings of the IEEE conference on computer vision and pattern recognition},
  pages={6809--6817},
  year={2017}
}

@article{liu2024visual,
  title={Visual instruction tuning},
  author={Liu, Haotian and Li, Chunyuan and Wu, Qingyang and Lee, Yong Jae},
  journal={Advances in neural information processing systems},
  volume={36},
  year={2024}
}

@INPROCEEDINGS{fathullah2024prompting,
  author={Fathullah, Yassir and Wu, Chunyang and Lakomkin, Egor and Jia, Junteng and Shangguan, Yuan and Li, Ke and Guo, Jinxi and Xiong, Wenhan and Mahadeokar, Jay and Kalinli, Ozlem and Fuegen, Christian and Seltzer, Mike},
  booktitle={ICASSP 2024 - 2024 IEEE International Conference on Acoustics, Speech and Signal Processing (ICASSP)}, 
  title={Prompting Large Language Models with Speech Recognition Abilities}, 
  year={2024},
  volume={},
  number={},
  pages={13351-13355},
  doi={10.1109/ICASSP48485.2024.10447605}}

@article{team2023gemini,
  title={Gemini: a family of highly capable multimodal models},
  author={Team, Gemini and Anil, Rohan and Borgeaud, Sebastian and Wu, Yonghui and Alayrac, Jean-Baptiste and Yu, Jiahui and Soricut, Radu and Schalkwyk, Johan and Dai, Andrew M and Hauth, Anja and others},
  journal={arXiv preprint arXiv:2312.11805},
  year={2023}
}

@article{touvron2023llama,
  title={Llama: Open and efficient foundation language models},
  author={Touvron, Hugo and Lavril, Thibaut and Izacard, Gautier and Martinet, Xavier and Lachaux, Marie-Anne and Lacroix, Timoth{\'e}e and Rozi{\`e}re, Baptiste and Goyal, Naman and Hambro, Eric and Azhar, Faisal and others},
  journal={arXiv preprint arXiv:2302.13971},
  year={2023}
}

@article{jiang2023mistral,
      title={Mistral 7B}, 
      author={Albert Q. Jiang and Alexandre Sablayrolles and Arthur Mensch and Chris Bamford and Devendra Singh Chaplot and Diego de las Casas and Florian Bressand and Gianna Lengyel and Guillaume Lample and Lucile Saulnier and Lélio Renard Lavaud and Marie-Anne Lachaux and Pierre Stock and Teven Le Scao and Thibaut Lavril and Thomas Wang and Timothée Lacroix and William El Sayed},
      year={2023},
      eprint={2310.06825},
      archivePrefix={arXiv},
      primaryClass={cs.CL},
      url={https://arxiv.org/abs/2310.06825}, 
}

@article{gunasekar2023textbooks,
  title={Textbooks are all you need},
  author={Gunasekar, Suriya and Zhang, Yi and Aneja, Jyoti and Mendes, Caio C{\'e}sar Teodoro and Del Giorno, Allie and Gopi, Sivakanth and Javaheripi, Mojan and Kauffmann, Piero and de Rosa, Gustavo and Saarikivi, Olli and others},
  journal={arXiv preprint arXiv:2306.11644},
  year={2023}
}

@article{tang2023semantic,
  title={Semantic reconstruction of continuous language from non-invasive brain recordings},
  author={Tang, Jerry and LeBel, Amanda and Jain, Shailee and Huth, Alexander G},
  journal={Nature Neuroscience},
  volume={26},
  number={5},
  pages={858--866},
  year={2023},
  publisher={Nature Publishing Group US New York}
}

@article{Speier_2016,
doi = {10.1088/1741-2560/13/3/031002},
url = {https://dx.doi.org/10.1088/1741-2560/13/3/031002},
year = {2016},
month = {may},
publisher = {IOP Publishing},
volume = {13},
number = {3},
pages = {031002},
author = {W Speier and C Arnold and N Pouratian},
title = {Integrating language models into classifiers for BCI communication: a review},
journal = {Journal of Neural Engineering}
}

@misc{achiam2023gpt,
      title={GPT-4 Technical Report}, 
      author={OpenAI and Josh Achiam and Steven Adler and Sandhini Agarwal and Lama Ahmad and Ilge Akkaya and Florencia Leoni Aleman and Diogo Almeida and Janko Altenschmidt and Sam Altman and Shyamal Anadkat and Red Avila and Igor Babuschkin and Suchir Balaji and Valerie Balcom and Paul Baltescu and Haiming Bao and Mohammad Bavarian and Jeff Belgum and Irwan Bello and Jake Berdine and Gabriel Bernadett-Shapiro and Christopher Berner and Lenny Bogdonoff and Oleg Boiko and Madelaine Boyd and Anna-Luisa Brakman and Greg Brockman and Tim Brooks and Miles Brundage and Kevin Button and Trevor Cai and Rosie Campbell and Andrew Cann and Brittany Carey and Chelsea Carlson and Rory Carmichael and Brooke Chan and Che Chang and Fotis Chantzis and Derek Chen and Sully Chen and Ruby Chen and Jason Chen and Mark Chen and Ben Chess and Chester Cho and Casey Chu and Hyung Won Chung and Dave Cummings and Jeremiah Currier and Yunxing Dai and Cory Decareaux and Thomas Degry and Noah Deutsch and Damien Deville and Arka Dhar and David Dohan and Steve Dowling and Sheila Dunning and Adrien Ecoffet and Atty Eleti and Tyna Eloundou and David Farhi and Liam Fedus and Niko Felix and Simón Posada Fishman and Juston Forte and Isabella Fulford and Leo Gao and Elie Georges and Christian Gibson and Vik Goel and Tarun Gogineni and Gabriel Goh and Rapha Gontijo-Lopes and Jonathan Gordon and Morgan Grafstein and Scott Gray and Ryan Greene and Joshua Gross and Shixiang Shane Gu and Yufei Guo and Chris Hallacy and Jesse Han and Jeff Harris and Yuchen He and Mike Heaton and Johannes Heidecke and Chris Hesse and Alan Hickey and Wade Hickey and Peter Hoeschele and Brandon Houghton and Kenny Hsu and Shengli Hu and Xin Hu and Joost Huizinga and Shantanu Jain and Shawn Jain and Joanne Jang and Angela Jiang and Roger Jiang and Haozhun Jin and Denny Jin and Shino Jomoto and Billie Jonn and Heewoo Jun and Tomer Kaftan and Łukasz Kaiser and Ali Kamali and Ingmar Kanitscheider and Nitish Shirish Keskar and Tabarak Khan and Logan Kilpatrick and Jong Wook Kim and Christina Kim and Yongjik Kim and Jan Hendrik Kirchner and Jamie Kiros and Matt Knight and Daniel Kokotajlo and Łukasz Kondraciuk and Andrew Kondrich and Aris Konstantinidis and Kyle Kosic and Gretchen Krueger and Vishal Kuo and Michael Lampe and Ikai Lan and Teddy Lee and Jan Leike and Jade Leung and Daniel Levy and Chak Ming Li and Rachel Lim and Molly Lin and Stephanie Lin and Mateusz Litwin and Theresa Lopez and Ryan Lowe and Patricia Lue and Anna Makanju and Kim Malfacini and Sam Manning and Todor Markov and Yaniv Markovski and Bianca Martin and Katie Mayer and Andrew Mayne and Bob McGrew and Scott Mayer McKinney and Christine McLeavey and Paul McMillan and Jake McNeil and David Medina and Aalok Mehta and Jacob Menick and Luke Metz and Andrey Mishchenko and Pamela Mishkin and Vinnie Monaco and Evan Morikawa and Daniel Mossing and Tong Mu and Mira Murati and Oleg Murk and David Mély and Ashvin Nair and Reiichiro Nakano and Rajeev Nayak and Arvind Neelakantan and Richard Ngo and Hyeonwoo Noh and Long Ouyang and Cullen O'Keefe and Jakub Pachocki and Alex Paino and Joe Palermo and Ashley Pantuliano and Giambattista Parascandolo and Joel Parish and Emy Parparita and Alex Passos and Mikhail Pavlov and Andrew Peng and Adam Perelman and Filipe de Avila Belbute Peres and Michael Petrov and Henrique Ponde de Oliveira Pinto and Michael and Pokorny and Michelle Pokrass and Vitchyr H. Pong and Tolly Powell and Alethea Power and Boris Power and Elizabeth Proehl and Raul Puri and Alec Radford and Jack Rae and Aditya Ramesh and Cameron Raymond and Francis Real and Kendra Rimbach and Carl Ross and Bob Rotsted and Henri Roussez and Nick Ryder and Mario Saltarelli and Ted Sanders and Shibani Santurkar and Girish Sastry and Heather Schmidt and David Schnurr and John Schulman and Daniel Selsam and Kyla Sheppard and Toki Sherbakov and Jessica Shieh and Sarah Shoker and Pranav Shyam and Szymon Sidor and Eric Sigler and Maddie Simens and Jordan Sitkin and Katarina Slama and Ian Sohl and Benjamin Sokolowsky and Yang Song and Natalie Staudacher and Felipe Petroski Such and Natalie Summers and Ilya Sutskever and Jie Tang and Nikolas Tezak and Madeleine B. Thompson and Phil Tillet and Amin Tootoonchian and Elizabeth Tseng and Preston Tuggle and Nick Turley and Jerry Tworek and Juan Felipe Cerón Uribe and Andrea Vallone and Arun Vijayvergiya and Chelsea Voss and Carroll Wainwright and Justin Jay Wang and Alvin Wang and Ben Wang and Jonathan Ward and Jason Wei and CJ Weinmann and Akila Welihinda and Peter Welinder and Jiayi Weng and Lilian Weng and Matt Wiethoff and Dave Willner and Clemens Winter and Samuel Wolrich and Hannah Wong and Lauren Workman and Sherwin Wu and Jeff Wu and Michael Wu and Kai Xiao and Tao Xu and Sarah Yoo and Kevin Yu and Qiming Yuan and Wojciech Zaremba and Rowan Zellers and Chong Zhang and Marvin Zhang and Shengjia Zhao and Tianhao Zheng and Juntang Zhuang and William Zhuk and Barret Zoph},
      year={2024},
      eprint={2303.08774},
      archivePrefix={arXiv},
      primaryClass={cs.CL},
      url={https://arxiv.org/abs/2303.08774}, 
}

@article{he2015noninvasive,
  title={Noninvasive brain-computer interfaces based on sensorimotor rhythms},
  author={He, Bin and Baxter, Bryan and Edelman, Bradley J and Cline, Christopher C and Wenjing, W Ye},
  journal={Proceedings of the IEEE},
  volume={103},
  number={6},
  pages={907--925},
  year={2015},
  publisher={IEEE}
}

@inproceedings{
benchetrit2023brain,
title={Brain decoding: toward real-time reconstruction of visual perception},
author={Yohann Benchetrit and Hubert Banville and Jean-Remi King},
booktitle={The Twelfth International Conference on Learning Representations},
year={2024},
url={https://openreview.net/forum?id=3y1K6buO8c}
}

@article{defossez2023decoding,
  title={Decoding speech perception from non-invasive brain recordings},
  author={D{\'e}fossez, Alexandre and Caucheteux, Charlotte and Rapin, J{\'e}r{\'e}my and Kabeli, Ori and King, Jean-R{\'e}mi},
  journal={Nature Machine Intelligence},
  volume={5},
  number={10},
  pages={1097--1107},
  year={2023},
  publisher={Nature Publishing Group UK London}
}

@article{rubenstein2023audiopalm,
  title={Audiopalm: A large language model that can speak and listen},
  author={Rubenstein, Paul K and Asawaroengchai, Chulayuth and Nguyen, Duc Dung and Bapna, Ankur and Borsos, Zal{\'a}n and Quitry, F{\'e}lix de Chaumont and Chen, Peter and Badawy, Dalia El and Han, Wei and Kharitonov, Eugene and others},
  journal={arXiv preprint arXiv:2306.12925},
  year={2023}
}

@misc{openai2024gpt4ocard,
      title={GPT-4o System Card}, 
      author={OpenAI and : and Aaron Hurst and Adam Lerer and Adam P. Goucher and Adam Perelman and Aditya Ramesh and Aidan Clark and AJ Ostrow and Akila Welihinda and Alan Hayes and Alec Radford and Aleksander Mądry and Alex Baker-Whitcomb and Alex Beutel and Alex Borzunov and Alex Carney and Alex Chow and Alex Kirillov and Alex Nichol and Alex Paino and Alex Renzin and Alex Tachard Passos and Alexander Kirillov and Alexi Christakis and Alexis Conneau and Ali Kamali and Allan Jabri and Allison Moyer and Allison Tam and Amadou Crookes and Amin Tootoochian and Amin Tootoonchian and Ananya Kumar and Andrea Vallone and Andrej Karpathy and Andrew Braunstein and Andrew Cann and Andrew Codispoti and Andrew Galu and Andrew Kondrich and Andrew Tulloch and Andrey Mishchenko and Angela Baek and Angela Jiang and Antoine Pelisse and Antonia Woodford and Anuj Gosalia and Arka Dhar and Ashley Pantuliano and Avi Nayak and Avital Oliver and Barret Zoph and Behrooz Ghorbani and Ben Leimberger and Ben Rossen and Ben Sokolowsky and Ben Wang and Benjamin Zweig and Beth Hoover and Blake Samic and Bob McGrew and Bobby Spero and Bogo Giertler and Bowen Cheng and Brad Lightcap and Brandon Walkin and Brendan Quinn and Brian Guarraci and Brian Hsu and Bright Kellogg and Brydon Eastman and Camillo Lugaresi and Carroll Wainwright and Cary Bassin and Cary Hudson and Casey Chu and Chad Nelson and Chak Li and Chan Jun Shern and Channing Conger and Charlotte Barette and Chelsea Voss and Chen Ding and Cheng Lu and Chong Zhang and Chris Beaumont and Chris Hallacy and Chris Koch and Christian Gibson and Christina Kim and Christine Choi and Christine McLeavey and Christopher Hesse and Claudia Fischer and Clemens Winter and Coley Czarnecki and Colin Jarvis and Colin Wei and Constantin Koumouzelis and Dane Sherburn and Daniel Kappler and Daniel Levin and Daniel Levy and David Carr and David Farhi and David Mely and David Robinson and David Sasaki and Denny Jin and Dev Valladares and Dimitris Tsipras and Doug Li and Duc Phong Nguyen and Duncan Findlay and Edede Oiwoh and Edmund Wong and Ehsan Asdar and Elizabeth Proehl and Elizabeth Yang and Eric Antonow and Eric Kramer and Eric Peterson and Eric Sigler and Eric Wallace and Eugene Brevdo and Evan Mays and Farzad Khorasani and Felipe Petroski Such and Filippo Raso and Francis Zhang and Fred von Lohmann and Freddie Sulit and Gabriel Goh and Gene Oden and Geoff Salmon and Giulio Starace and Greg Brockman and Hadi Salman and Haiming Bao and Haitang Hu and Hannah Wong and Haoyu Wang and Heather Schmidt and Heather Whitney and Heewoo Jun and Hendrik Kirchner and Henrique Ponde de Oliveira Pinto and Hongyu Ren and Huiwen Chang and Hyung Won Chung and Ian Kivlichan and Ian O'Connell and Ian O'Connell and Ian Osband and Ian Silber and Ian Sohl and Ibrahim Okuyucu and Ikai Lan and Ilya Kostrikov and Ilya Sutskever and Ingmar Kanitscheider and Ishaan Gulrajani and Jacob Coxon and Jacob Menick and Jakub Pachocki and James Aung and James Betker and James Crooks and James Lennon and Jamie Kiros and Jan Leike and Jane Park and Jason Kwon and Jason Phang and Jason Teplitz and Jason Wei and Jason Wolfe and Jay Chen and Jeff Harris and Jenia Varavva and Jessica Gan Lee and Jessica Shieh and Ji Lin and Jiahui Yu and Jiayi Weng and Jie Tang and Jieqi Yu and Joanne Jang and Joaquin Quinonero Candela and Joe Beutler and Joe Landers and Joel Parish and Johannes Heidecke and John Schulman and Jonathan Lachman and Jonathan McKay and Jonathan Uesato and Jonathan Ward and Jong Wook Kim and Joost Huizinga and Jordan Sitkin and Jos Kraaijeveld and Josh Gross and Josh Kaplan and Josh Snyder and Joshua Achiam and Joy Jiao and Joyce Lee and Juntang Zhuang and Justyn Harriman and Kai Fricke and Kai Hayashi and Karan Singhal and Katy Shi and Kavin Karthik and Kayla Wood and Kendra Rimbach and Kenny Hsu and Kenny Nguyen and Keren Gu-Lemberg and Kevin Button and Kevin Liu and Kiel Howe and Krithika Muthukumar and Kyle Luther and Lama Ahmad and Larry Kai and Lauren Itow and Lauren Workman and Leher Pathak and Leo Chen and Li Jing and Lia Guy and Liam Fedus and Liang Zhou and Lien Mamitsuka and Lilian Weng and Lindsay McCallum and Lindsey Held and Long Ouyang and Louis Feuvrier and Lu Zhang and Lukas Kondraciuk and Lukasz Kaiser and Luke Hewitt and Luke Metz and Lyric Doshi and Mada Aflak and Maddie Simens and Madelaine Boyd and Madeleine Thompson and Marat Dukhan and Mark Chen and Mark Gray and Mark Hudnall and Marvin Zhang and Marwan Aljubeh and Mateusz Litwin and Matthew Zeng and Max Johnson and Maya Shetty and Mayank Gupta and Meghan Shah and Mehmet Yatbaz and Meng Jia Yang and Mengchao Zhong and Mia Glaese and Mianna Chen and Michael Janner and Michael Lampe and Michael Petrov and Michael Wu and Michele Wang and Michelle Fradin and Michelle Pokrass and Miguel Castro and Miguel Oom Temudo de Castro and Mikhail Pavlov and Miles Brundage and Miles Wang and Minal Khan and Mira Murati and Mo Bavarian and Molly Lin and Murat Yesildal and Nacho Soto and Natalia Gimelshein and Natalie Cone and Natalie Staudacher and Natalie Summers and Natan LaFontaine and Neil Chowdhury and Nick Ryder and Nick Stathas and Nick Turley and Nik Tezak and Niko Felix and Nithanth Kudige and Nitish Keskar and Noah Deutsch and Noel Bundick and Nora Puckett and Ofir Nachum and Ola Okelola and Oleg Boiko and Oleg Murk and Oliver Jaffe and Olivia Watkins and Olivier Godement and Owen Campbell-Moore and Patrick Chao and Paul McMillan and Pavel Belov and Peng Su and Peter Bak and Peter Bakkum and Peter Deng and Peter Dolan and Peter Hoeschele and Peter Welinder and Phil Tillet and Philip Pronin and Philippe Tillet and Prafulla Dhariwal and Qiming Yuan and Rachel Dias and Rachel Lim and Rahul Arora and Rajan Troll and Randall Lin and Rapha Gontijo Lopes and Raul Puri and Reah Miyara and Reimar Leike and Renaud Gaubert and Reza Zamani and Ricky Wang and Rob Donnelly and Rob Honsby and Rocky Smith and Rohan Sahai and Rohit Ramchandani and Romain Huet and Rory Carmichael and Rowan Zellers and Roy Chen and Ruby Chen and Ruslan Nigmatullin and Ryan Cheu and Saachi Jain and Sam Altman and Sam Schoenholz and Sam Toizer and Samuel Miserendino and Sandhini Agarwal and Sara Culver and Scott Ethersmith and Scott Gray and Sean Grove and Sean Metzger and Shamez Hermani and Shantanu Jain and Shengjia Zhao and Sherwin Wu and Shino Jomoto and Shirong Wu and Shuaiqi and Xia and Sonia Phene and Spencer Papay and Srinivas Narayanan and Steve Coffey and Steve Lee and Stewart Hall and Suchir Balaji and Tal Broda and Tal Stramer and Tao Xu and Tarun Gogineni and Taya Christianson and Ted Sanders and Tejal Patwardhan and Thomas Cunninghman and Thomas Degry and Thomas Dimson and Thomas Raoux and Thomas Shadwell and Tianhao Zheng and Todd Underwood and Todor Markov and Toki Sherbakov and Tom Rubin and Tom Stasi and Tomer Kaftan and Tristan Heywood and Troy Peterson and Tyce Walters and Tyna Eloundou and Valerie Qi and Veit Moeller and Vinnie Monaco and Vishal Kuo and Vlad Fomenko and Wayne Chang and Weiyi Zheng and Wenda Zhou and Wesam Manassra and Will Sheu and Wojciech Zaremba and Yash Patil and Yilei Qian and Yongjik Kim and Youlong Cheng and Yu Zhang and Yuchen He and Yuchen Zhang and Yujia Jin and Yunxing Dai and Yury Malkov},
      year={2024},
      eprint={2410.21276},
      archivePrefix={arXiv},
      primaryClass={cs.CL},
      url={https://arxiv.org/abs/2410.21276}, 
}

@misc{grattafiori2024llama3herdmodels,
      title={The Llama 3 Herd of Models}, 
      author={Aaron Grattafiori and Abhimanyu Dubey and Abhinav Jauhri and Abhinav Pandey and Abhishek Kadian and Ahmad Al-Dahle and Aiesha Letman and Akhil Mathur and Alan Schelten and Alex Vaughan and Amy Yang and Angela Fan and Anirudh Goyal and Anthony Hartshorn and Aobo Yang and Archi Mitra and Archie Sravankumar and Artem Korenev and Arthur Hinsvark and Arun Rao and Aston Zhang and Aurelien Rodriguez and Austen Gregerson and Ava Spataru and Baptiste Roziere and Bethany Biron and Binh Tang and Bobbie Chern and Charlotte Caucheteux and Chaya Nayak and Chloe Bi and Chris Marra and Chris McConnell and Christian Keller and Christophe Touret and Chunyang Wu and Corinne Wong and Cristian Canton Ferrer and Cyrus Nikolaidis and Damien Allonsius and Daniel Song and Danielle Pintz and Danny Livshits and Danny Wyatt and David Esiobu and Dhruv Choudhary and Dhruv Mahajan and Diego Garcia-Olano and Diego Perino and Dieuwke Hupkes and Egor Lakomkin and Ehab AlBadawy and Elina Lobanova and Emily Dinan and Eric Michael Smith and Filip Radenovic and Francisco Guzmán and Frank Zhang and Gabriel Synnaeve and Gabrielle Lee and Georgia Lewis Anderson and Govind Thattai and Graeme Nail and Gregoire Mialon and Guan Pang and Guillem Cucurell and Hailey Nguyen and Hannah Korevaar and Hu Xu and Hugo Touvron and Iliyan Zarov and Imanol Arrieta Ibarra and Isabel Kloumann and Ishan Misra and Ivan Evtimov and Jack Zhang and Jade Copet and Jaewon Lee and Jan Geffert and Jana Vranes and Jason Park and Jay Mahadeokar and Jeet Shah and Jelmer van der Linde and Jennifer Billock and Jenny Hong and Jenya Lee and Jeremy Fu and Jianfeng Chi and Jianyu Huang and Jiawen Liu and Jie Wang and Jiecao Yu and Joanna Bitton and Joe Spisak and Jongsoo Park and Joseph Rocca and Joshua Johnstun and Joshua Saxe and Junteng Jia and Kalyan Vasuden Alwala and Karthik Prasad and Kartikeya Upasani and Kate Plawiak and Ke Li and Kenneth Heafield and Kevin Stone and Khalid El-Arini and Krithika Iyer and Kshitiz Malik and Kuenley Chiu and Kunal Bhalla and Kushal Lakhotia and Lauren Rantala-Yeary and Laurens van der Maaten and Lawrence Chen and Liang Tan and Liz Jenkins and Louis Martin and Lovish Madaan and Lubo Malo and Lukas Blecher and Lukas Landzaat and Luke de Oliveira and Madeline Muzzi and Mahesh Pasupuleti and Mannat Singh and Manohar Paluri and Marcin Kardas and Maria Tsimpoukelli and Mathew Oldham and Mathieu Rita and Maya Pavlova and Melanie Kambadur and Mike Lewis and Min Si and Mitesh Kumar Singh and Mona Hassan and Naman Goyal and Narjes Torabi and Nikolay Bashlykov and Nikolay Bogoychev and Niladri Chatterji and Ning Zhang and Olivier Duchenne and Onur Çelebi and Patrick Alrassy and Pengchuan Zhang and Pengwei Li and Petar Vasic and Peter Weng and Prajjwal Bhargava and Pratik Dubal and Praveen Krishnan and Punit Singh Koura and Puxin Xu and Qing He and Qingxiao Dong and Ragavan Srinivasan and Raj Ganapathy and Ramon Calderer and Ricardo Silveira Cabral and Robert Stojnic and Roberta Raileanu and Rohan Maheswari and Rohit Girdhar and Rohit Patel and Romain Sauvestre and Ronnie Polidoro and Roshan Sumbaly and Ross Taylor and Ruan Silva and Rui Hou and Rui Wang and Saghar Hosseini and Sahana Chennabasappa and Sanjay Singh and Sean Bell and Seohyun Sonia Kim and Sergey Edunov and Shaoliang Nie and Sharan Narang and Sharath Raparthy and Sheng Shen and Shengye Wan and Shruti Bhosale and Shun Zhang and Simon Vandenhende and Soumya Batra and Spencer Whitman and Sten Sootla and Stephane Collot and Suchin Gururangan and Sydney Borodinsky and Tamar Herman and Tara Fowler and Tarek Sheasha and Thomas Georgiou and Thomas Scialom and Tobias Speckbacher and Todor Mihaylov and Tong Xiao and Ujjwal Karn and Vedanuj Goswami and Vibhor Gupta and Vignesh Ramanathan and Viktor Kerkez and Vincent Gonguet and Virginie Do and Vish Vogeti and Vítor Albiero and Vladan Petrovic and Weiwei Chu and Wenhan Xiong and Wenyin Fu and Whitney Meers and Xavier Martinet and Xiaodong Wang and Xiaofang Wang and Xiaoqing Ellen Tan and Xide Xia and Xinfeng Xie and Xuchao Jia and Xuewei Wang and Yaelle Goldschlag and Yashesh Gaur and Yasmine Babaei and Yi Wen and Yiwen Song and Yuchen Zhang and Yue Li and Yuning Mao and Zacharie Delpierre Coudert and Zheng Yan and Zhengxing Chen and Zoe Papakipos and Aaditya Singh and Aayushi Srivastava and Abha Jain and Adam Kelsey and Adam Shajnfeld and Adithya Gangidi and Adolfo Victoria and Ahuva Goldstand and Ajay Menon and Ajay Sharma and Alex Boesenberg and Alexei Baevski and Allie Feinstein and Amanda Kallet and Amit Sangani and Amos Teo and Anam Yunus and Andrei Lupu and Andres Alvarado and Andrew Caples and Andrew Gu and Andrew Ho and Andrew Poulton and Andrew Ryan and Ankit Ramchandani and Annie Dong and Annie Franco and Anuj Goyal and Aparajita Saraf and Arkabandhu Chowdhury and Ashley Gabriel and Ashwin Bharambe and Assaf Eisenman and Azadeh Yazdan and Beau James and Ben Maurer and Benjamin Leonhardi and Bernie Huang and Beth Loyd and Beto De Paola and Bhargavi Paranjape and Bing Liu and Bo Wu and Boyu Ni and Braden Hancock and Bram Wasti and Brandon Spence and Brani Stojkovic and Brian Gamido and Britt Montalvo and Carl Parker and Carly Burton and Catalina Mejia and Ce Liu and Changhan Wang and Changkyu Kim and Chao Zhou and Chester Hu and Ching-Hsiang Chu and Chris Cai and Chris Tindal and Christoph Feichtenhofer and Cynthia Gao and Damon Civin and Dana Beaty and Daniel Kreymer and Daniel Li and David Adkins and David Xu and Davide Testuggine and Delia David and Devi Parikh and Diana Liskovich and Didem Foss and Dingkang Wang and Duc Le and Dustin Holland and Edward Dowling and Eissa Jamil and Elaine Montgomery and Eleonora Presani and Emily Hahn and Emily Wood and Eric-Tuan Le and Erik Brinkman and Esteban Arcaute and Evan Dunbar and Evan Smothers and Fei Sun and Felix Kreuk and Feng Tian and Filippos Kokkinos and Firat Ozgenel and Francesco Caggioni and Frank Kanayet and Frank Seide and Gabriela Medina Florez and Gabriella Schwarz and Gada Badeer and Georgia Swee and Gil Halpern and Grant Herman and Grigory Sizov and Guangyi and Zhang and Guna Lakshminarayanan and Hakan Inan and Hamid Shojanazeri and Han Zou and Hannah Wang and Hanwen Zha and Haroun Habeeb and Harrison Rudolph and Helen Suk and Henry Aspegren and Hunter Goldman and Hongyuan Zhan and Ibrahim Damlaj and Igor Molybog and Igor Tufanov and Ilias Leontiadis and Irina-Elena Veliche and Itai Gat and Jake Weissman and James Geboski and James Kohli and Janice Lam and Japhet Asher and Jean-Baptiste Gaya and Jeff Marcus and Jeff Tang and Jennifer Chan and Jenny Zhen and Jeremy Reizenstein and Jeremy Teboul and Jessica Zhong and Jian Jin and Jingyi Yang and Joe Cummings and Jon Carvill and Jon Shepard and Jonathan McPhie and Jonathan Torres and Josh Ginsburg and Junjie Wang and Kai Wu and Kam Hou U and Karan Saxena and Kartikay Khandelwal and Katayoun Zand and Kathy Matosich and Kaushik Veeraraghavan and Kelly Michelena and Keqian Li and Kiran Jagadeesh and Kun Huang and Kunal Chawla and Kyle Huang and Lailin Chen and Lakshya Garg and Lavender A and Leandro Silva and Lee Bell and Lei Zhang and Liangpeng Guo and Licheng Yu and Liron Moshkovich and Luca Wehrstedt and Madian Khabsa and Manav Avalani and Manish Bhatt and Martynas Mankus and Matan Hasson and Matthew Lennie and Matthias Reso and Maxim Groshev and Maxim Naumov and Maya Lathi and Meghan Keneally and Miao Liu and Michael L. Seltzer and Michal Valko and Michelle Restrepo and Mihir Patel and Mik Vyatskov and Mikayel Samvelyan and Mike Clark and Mike Macey and Mike Wang and Miquel Jubert Hermoso and Mo Metanat and Mohammad Rastegari and Munish Bansal and Nandhini Santhanam and Natascha Parks and Natasha White and Navyata Bawa and Nayan Singhal and Nick Egebo and Nicolas Usunier and Nikhil Mehta and Nikolay Pavlovich Laptev and Ning Dong and Norman Cheng and Oleg Chernoguz and Olivia Hart and Omkar Salpekar and Ozlem Kalinli and Parkin Kent and Parth Parekh and Paul Saab and Pavan Balaji and Pedro Rittner and Philip Bontrager and Pierre Roux and Piotr Dollar and Polina Zvyagina and Prashant Ratanchandani and Pritish Yuvraj and Qian Liang and Rachad Alao and Rachel Rodriguez and Rafi Ayub and Raghotham Murthy and Raghu Nayani and Rahul Mitra and Rangaprabhu Parthasarathy and Raymond Li and Rebekkah Hogan and Robin Battey and Rocky Wang and Russ Howes and Ruty Rinott and Sachin Mehta and Sachin Siby and Sai Jayesh Bondu and Samyak Datta and Sara Chugh and Sara Hunt and Sargun Dhillon and Sasha Sidorov and Satadru Pan and Saurabh Mahajan and Saurabh Verma and Seiji Yamamoto and Sharadh Ramaswamy and Shaun Lindsay and Shaun Lindsay and Sheng Feng and Shenghao Lin and Shengxin Cindy Zha and Shishir Patil and Shiva Shankar and Shuqiang Zhang and Shuqiang Zhang and Sinong Wang and Sneha Agarwal and Soji Sajuyigbe and Soumith Chintala and Stephanie Max and Stephen Chen and Steve Kehoe and Steve Satterfield and Sudarshan Govindaprasad and Sumit Gupta and Summer Deng and Sungmin Cho and Sunny Virk and Suraj Subramanian and Sy Choudhury and Sydney Goldman and Tal Remez and Tamar Glaser and Tamara Best and Thilo Koehler and Thomas Robinson and Tianhe Li and Tianjun Zhang and Tim Matthews and Timothy Chou and Tzook Shaked and Varun Vontimitta and Victoria Ajayi and Victoria Montanez and Vijai Mohan and Vinay Satish Kumar and Vishal Mangla and Vlad Ionescu and Vlad Poenaru and Vlad Tiberiu Mihailescu and Vladimir Ivanov and Wei Li and Wenchen Wang and Wenwen Jiang and Wes Bouaziz and Will Constable and Xiaocheng Tang and Xiaojian Wu and Xiaolan Wang and Xilun Wu and Xinbo Gao and Yaniv Kleinman and Yanjun Chen and Ye Hu and Ye Jia and Ye Qi and Yenda Li and Yilin Zhang and Ying Zhang and Yossi Adi and Youngjin Nam and Yu and Wang and Yu Zhao and Yuchen Hao and Yundi Qian and Yunlu Li and Yuzi He and Zach Rait and Zachary DeVito and Zef Rosnbrick and Zhaoduo Wen and Zhenyu Yang and Zhiwei Zhao and Zhiyu Ma},
      year={2024},
      eprint={2407.21783},
      archivePrefix={arXiv},
      primaryClass={cs.AI},
      url={https://arxiv.org/abs/2407.21783}, 
}

@misc{qwen2025qwen25technicalreport,
      title={Qwen2.5 Technical Report}, 
      author={Qwen and : and An Yang and Baosong Yang and Beichen Zhang and Binyuan Hui and Bo Zheng and Bowen Yu and Chengyuan Li and Dayiheng Liu and Fei Huang and Haoran Wei and Huan Lin and Jian Yang and Jianhong Tu and Jianwei Zhang and Jianxin Yang and Jiaxi Yang and Jingren Zhou and Junyang Lin and Kai Dang and Keming Lu and Keqin Bao and Kexin Yang and Le Yu and Mei Li and Mingfeng Xue and Pei Zhang and Qin Zhu and Rui Men and Runji Lin and Tianhao Li and Tianyi Tang and Tingyu Xia and Xingzhang Ren and Xuancheng Ren and Yang Fan and Yang Su and Yichang Zhang and Yu Wan and Yuqiong Liu and Zeyu Cui and Zhenru Zhang and Zihan Qiu},
      year={2025},
      eprint={2412.15115},
      archivePrefix={arXiv},
      primaryClass={cs.CL},
      url={https://arxiv.org/abs/2412.15115}, 
}

@article{lamprou2025creating,
  title={On creating a brain-to-text decoder},
  author={Lamprou, Zenon and Moshfeghi, Yashar},
  journal={arXiv preprint arXiv:2501.06326},
  year={2025}
}

@inproceedings{wang2024enhancing,
  title={Enhancing EEG-to-text decoding through transferable representations from pre-trained contrastive EEG-text masked autoencoder},
  author={Wang, Jiaqi and Song, Zhenxi and Ma, Zhengyu and Qiu, Xipeng and Zhang, Min and Zhang, Zhiguo},
  booktitle={Proceedings of the 62nd Annual Meeting of the Association for Computational Linguistics (Volume 1: Long Papers)},
  pages={7278--7292},
  year={2024}
}

@article{liu2025learning,
  title={Learning Interpretable Representations Leads to Semantically Faithful EEG-to-Text Generation},
  author={Liu, Xiaozhao and Shen, Dinggang and Liu, Xihui},
  journal={arXiv preprint arXiv:2505.17099},
  year={2025}
}

@InProceedings{pmlr-v139-radford21a,
  title = 	 {Learning Transferable Visual Models From Natural Language Supervision},
  author =       {Radford, Alec and Kim, Jong Wook and Hallacy, Chris and Ramesh, Aditya and Goh, Gabriel and Agarwal, Sandhini and Sastry, Girish and Askell, Amanda and Mishkin, Pamela and Clark, Jack and Krueger, Gretchen and Sutskever, Ilya},
  booktitle = 	 {Proceedings of the 38th International Conference on Machine Learning},
  pages = 	 {8748--8763},
  year = 	 {2021},
  editor = 	 {Meila, Marina and Zhang, Tong},
  volume = 	 {139},
  series = 	 {Proceedings of Machine Learning Research},
  month = 	 {18--24 Jul},
  publisher =    {PMLR},
  url = 	 {https://proceedings.mlr.press/v139/radford21a.html}
}

@article{willett2023high,
  title={A high-performance speech neuroprosthesis},
  author={Willett, Francis R and Kunz, Erin M and Fan, Chaofei and Avansino, Donald T and Wilson, Guy H and Choi, Eun Young and Kamdar, Foram and Glasser, Matthew F and Hochberg, Leigh R and Druckmann, Shaul and others},
  journal={Nature},
  volume={620},
  number={7976},
  pages={1031--1036},
  year={2023},
  publisher={Nature Publishing Group UK London}
}

@article{metzger2022generalizable,
  title={Generalizable spelling using a speech neuroprosthesis in an individual with severe limb and vocal paralysis},
  author={Metzger, Sean L and Liu, Jessie R and Moses, David A and Dougherty, Maximilian E and Seaton, Margaret P and Littlejohn, Kaylo T and Chartier, Josh and Anumanchipalli, Gopala K and Tu-Chan, Adelyn and Ganguly, Karunesh and others},
  journal={Nature communications},
  volume={13},
  number={1},
  pages={6510},
  year={2022},
  publisher={Nature Publishing Group UK London}
}

@article{miller1995wordnet,
  title={WordNet: a lexical database for English},
  author={Miller, George A},
  journal={Communications of the ACM},
  volume={38},
  number={11},
  pages={39--41},
  year={1995},
  publisher={ACM New York, NY, USA}
}





\appendix

\section{LLM Prompt Templates}

This appendix provides the structured zero-shot prompting templates used to translate EEG-derived semantic anchors into fluent natural language descriptions. These templates define the interface between the local on-device retrieval stage and the remote Large Language Model (LLM) decoder, ensuring that the synthesized text remains grounded in the authentic neural signal while maintaining syntactic cohesion. We detail both the standard configuration—which utilizes the primary object anchor—and the ablation configuration used to evaluate the generative capacity of the extracted Bag-of-Words (BoW) alone. 

\subsection{Standard Prompt (\textsc{WithObj})}
The following template is used for the primary decoding task, incorporating both the predicted object label and the extracted Bag-of-Words (BoW).

You are given an object label and a noisy bag-of-words (BoW). Both object label and BoW will be accompanied with numbers, the numbers with object labels are the softmax probabilities of correctly guessing the object label, and the BoW are cosine similarities of the words to our embedding.

Your goal is to regenerate the most likely original image caption.

Instructions: \\
- Use the object label as a possible anchor. \\
- Use the similarity scores to infer which words are relevant. \\
- Ignore or remove garbage, irrelevant, contradictory, or low-signal words. \\
- Use only a small, coherent subset of the BoW plus the object label. \\
- Do NOT invent new objects not supported by the label or high-similarity words. \\

Output:
Return ONLY one natural-language caption (8–20 words). No explanations, no lists, no formatting.

Input:
Object label: \{pred\_obj\} (prob: \{pred\_conf:.4f\})

BoW tokens with scores:
\{words\_str\}

\subsection{Ablation Prompt (\textsc{WithoutObj})}
The following template is used for the ablation study, where the language model reconstructs the caption using only the EEG-derived semantic anchors.

You are given a noisy bag-of-words (BoW). BoW will be accompanied with numbers, the numbers with BoW are cosine similarities of the words to our embedding.

Your goal is to regenerate the most likely original image caption.

Instructions: \\
- Use the similarity scores to infer which words are relevant. \\
- Ignore or remove garbage, irrelevant, contradictory, or low-signal words.\\
- Use only a small, coherent subset of the BoW. \\
- Do NOT invent new objects not supported by the high-similarity words.\\

Output:
Return ONLY one natural-language caption (8–20 words). No explanations, no lists, no formatting.

Input:
BoW tokens with scores:
\{words\_str\}


\end{document}